\documentclass[3p]{elsarticle}

\usepackage{algorithm,algorithmic}
\usepackage{amsmath}
\usepackage{amsfonts,amssymb}
\usepackage{multirow}
\usepackage{supertabular}
\usepackage{subfigure}
\usepackage{booktabs}
\usepackage{url}
\usepackage{float}
\journal{Journal of Information Sciences}
\begin{document}
 \begin{frontmatter}
\title{Disentangled Variational Auto-Encoder \\for Semi-supervised Learning}  
 
\author[npu]{Yang Li}
\author[npu]{Quan Pan}
\author[asu]{Suhang Wang}
\author[ntu]{Haiyun Peng}
\author[npu]{Tao Yang }
\author[ntu]{Erik Cambria\corref{mycorrespondingauthor}}  
\cortext[mycorrespondingauthor]{Erik Cambria}  
\ead{cambria@ntu.edu.sg}
 
\address[npu]{School of Automation, Northwestern Polytechnical University}
\address[ntu]{School of Computer Science and Engineering, Nanyang Technological University}
\address[asu]{College of Information Sciences and Technology, Pennsylvania State University}

\begin{abstract}
Semi-supervised learning is attracting increasing attention due to the fact that datasets of many domains lack enough labeled data. Variational Auto-Encoder (VAE), in particular, has demonstrated the benefits of semi-supervised learning. The majority of existing semi-supervised VAEs utilize a classifier to exploit label information, where the parameters of the classifier are introduced to the VAE. Given the limited labeled data, learning the parameters for the classifiers may not be an optimal solution for exploiting label information. Therefore, in this paper, we develop a novel approach for semi-supervised VAE without classifier. Specifically, we propose a new model called Semi-supervised Disentangled VAE (SDVAE), which encodes the input data into disentangled representation and non-interpretable representation, then the category information is directly utilized to regularize the disentangled representation via the equality constraint. To further enhance the feature learning ability of the proposed VAE, we incorporate reinforcement learning to relieve the lack of data. The dynamic framework is capable of dealing with both image and text data with its corresponding encoder and decoder networks. Extensive experiments on image and text datasets demonstrate the effectiveness of the proposed framework.
\end{abstract}
\begin{keyword}
 Semi-supervised Learning \sep Variational Auto-encoder \sep Disentangled Representation \sep Neural Networks
\end{keyword} 
\end{frontmatter}

\section{Introduction}
The abundant data generated online every day has greatly advanced machine learning, data mining and computer vision communities~\cite{yourec,wang2018deep}. However, manual labeling of the large dataset is very time- and labor-consuming. Sometimes, it even requires domain knowledge. The majority datasets are with limited label. Therefore, semi-supervised learning, which utilizes both labeled and unlabeled data for model training, is attracting increasing attention~\cite{hussem,dai2015semi,odena2016semi,kingma2014semi,xu2017variational}. Existing semi-supervised models can be generally categorized into three main categories, i.e., discriminative models, generative models, and graph-based models, plus combinations of such categories~\cite{vapnik1977structural,he2007graph,huang2014semi,peng2015enhanced,wang2016linked}. 

Among various semi-supervised models proposed, the semi-supervised generative models based on variational auto-encoder (VAE) have shown strong performance in image classification~\cite{kingma2014semi,maaloe2016auxiliary} and text classification~\cite{xu2017variational,liigen}. The effectiveness of VAE for semi-supervised learning comes from its efficiency in posterior distribution estimation and its powerful ability in extracting features from text data~\cite{bowman2015generating} and image data~\cite{kingma2014semi,maaloe2016auxiliary}. To adapt VAE for semi-supervised learning, the semi-supervised VAEs are typically composed of three main components: an encoder network $q_{\phi}(z |x, y)$, a decoder $p_{\theta}(x | y, z)$ and a classifier $q_{\phi} (y | x)$. In the application, the encoder, decoder, and classifier can be implemented using various models, e.g., Multilayer Perceptron (MLP) or Convolutional Neural Network (CNN)~\cite{maaloe2016auxiliary,yan2016attribute2image}. Though the classifier plays a vital role in achieving the semi-supervised goal, it introduces extra parameters of itself to learn. With the limited labeled data, it may not be an optimal choice to introduce more parameters to VAE for semi-supervised learning because it may memorize the limited data with large quantities of parameters, namely overfitting.

Therefore, in this paper, we investigate if we can directly incorporate the limited label information to VAE without introducing the non-linear classifier so as to achieve the goal of semi-supervised learning.  In particular, we investigate the following two challenges: (1) Without introducing a classifier, how do we incorporate the label information to VAE for semi-supervised learning? and (2) How can we effectively use the label information for representation learning of VAE? In an attempt to solve these two challenges, we propose a novel semi-supervised learning model named semi-supervised disentangled VAE (SDVAE). SDVAE adopts the VAE with Karush-Kuhn-Tucker (KKT) conditions~\cite{KKT} as it has better representation learning ability than VAE. Unlike existing semi-supervised VAEs that utilize classifiers, SDVAE encodes the input data into disentangled representation and non-interpretable representation, and the category information is directly utilized to regularize the disentangled representation as an equality constraint, and the classification results can be obtained from the disentangled representation directly. As the labeled data is limited, the labeled information may not affect the model much. To this end, we further change the equality constraint into the reinforcement learning format, which helps the objective gain the category information heuristics. The inverse auto-regression (IAF)~\cite{kingma2016improved} is also applied to improve the latent variable learning. The proposed framework is flexible in which it can deal with both image and text data by choosing corresponding encoder and decoder networks. The main contributions of the paper are:
\begin{itemize}
  \item{Propose a novel semi-supervised framework which directly exploits the label information to regularize disentangled representation with reinforcement learning;}
  \item{Extract the disentangled variable for classification and the non-interpretable variable for the reconstruction from the data directly; and}
  \item{Conduct extensive experiments on image and text datasets to demonstrate the effectiveness of the proposed SDVAE.}
\end{itemize}

\section{Related Works} \label{sec:related_works}
In this section, we briefly review related works. Our work is related to semi-supervised learning, semi-supervised VAE, and variants of VAEs.

\subsection{Semi-supervised learning} 
 Semi-supervised learning is attracting increasing attention, and a lot of works have been proposed in this field~\cite{hu2017toward,he2007graph,odena2016semi,wang2016linked,peng2015enhanced,kingma2014semi,subramanya2014graph,xu2017variational}. Those works can be generally divided into four categories, i.e., discriminative models~\cite{vapnik1977structural,dai2015semi,huang2014semi,joachims2006transductive}, generative models~\cite{xu2017variational,hu2017toward,odena2016semi,bowman2015generating,kingma2014semi}, graph-based models~\cite{subramanya2014graph,he2007graph}, and the combination of those~\cite{he2007graph,fujino2008semisupervised}. The discriminative models aim to train a classifier that is able to find the hyperplane that separates both the labeled data and the unlabeled data~\cite{joachims2006transductive,vapnik1977structural,huang2014semi}. The generative model tries to inference the posterior distribution based on the Bayesian theory, then the label of data can be estimated based on the generative distribution~\cite{bowman2015generating,odena2016semi,kingma2014semi}. The nature of the graph-based model is the label propagation. After calculating the distance between unlabeled data and labeled data, the label of unlabeled data can be decided by the nearest labeled data~\cite{subramanya2014graph,he2007graph,peterson2009k}. Many works combine two or three models from different categories so as to take their advantages~\cite{he2007graph,fujino2008semisupervised}. For example, He et al.~\cite{he2007graph} investigated the generative model under the graph-based framework~\cite{he2007graph}; Fujino et al.~\cite{fujino2008semisupervised} studied semi-supervised learning with hybrid generative/discriminative classifier based on the maximum entropy principle. Recently, semi-supervised learning based on VAE has shown promising performance and has attracted increasing attention. Next, we will introduce semi-supervised VAE.

\subsection{Semi-supervised VAE} 
 Because of the effectiveness of deep generative models in capturing data distribution, semi-supervised models based on deep generative models such as generative adversarial network~\cite{springenberg2015unsupervised} and VAE~\cite{kingma2014semi} are becoming very popular. Various semi-supervised models based on VAE are proposed~\cite{kingma2014semi,xu2017variational}. A typical VAE is composed of an encoder network $q_{\phi}(z |x)$ which encodes the input $x$ to latent representation $z$, and a decoder network $p_{\theta}(x | z)$ which reconstructs $x$ from $z$. The essential idea of semi-supervised VAE is to add a classifier on top of the latent representation. Thus, the semi-supervised VAEs are typically composed of three main components: an encoder network $q_{\phi}(z |x, y)$, a decoder $p_{\theta}(x | y, z)$ and a classifier $q_{\phi} (y | x)$. For example, Semi-VAE~\cite{kingma2014semi} incorporates learned latent variable into a classifier and improves the performance greatly. SSVAE~\cite{xu2017variational} extends Semi-VAE for sequence data and also demonstrates its effectiveness in the semi-supervised learning on the text data. The aforementioned semi-supervised VAE all use a parametric classifier, which increases the burden to learn more parameters given the limited labeled data. Therefore, in this work, the proposed framework incorporates the label information directly into the disentangled representation and thus avoids the parametric classifier.

\subsection{Variants of VAE} Because of the great potential of VAE in image and text mining, various models based on VAE are proposed to further improve its performance~\cite{kingma2013auto,higgins2017beta,hoffman2016elbo,kingma2016improved}. For example, Higgins et al.~\cite{higgins2017beta} apply the KKT condition in the VAE, which gives a tighter lower bound. Similarly, Burda et al.~\cite{burda2015importance} introduce importance weighting to VAE, which also tries to give a tighter bound. Pu et al.~\cite{pu2017stein} consider the stein based sampling to minimize the Kullback-Leibler divergence (KL). Hoffman et al.~\cite{hoffman2016elbo} rewrite the evidence lower bound (ELBO) objective by decomposition, and give a clear explanation of each term. To extend the flexible of posterior inference, IAF is introduced~\cite{kingma2016improved} which improves the VAE a lot.

In this paper, SDVAE belongs to the combination of generative model and discriminative model, after estimating the posterior distribution of the data, an equality constraint on disentangled representation is added to guide the learning of the latent representation, together with heuristic inference, SDVAE is more effective in semi-supervised learning both in text and image data.

\section{Preliminaries} \label{sec:preliminaries}
In this section, we introduce preliminaries that will be useful to understand our model, which mainly cover details of VAE, VAE with KKT condition and semi-supervised VAE.

\subsection{Variational Auto-Encoder}
VAEs hasve emerged as one of the most popular deep generative models. One key step of VAE is to evaluate $p_{\theta}(x)$, which can be interpreted as
 \begin{equation}
 \label{equ:original_equation}
 \begin{aligned}
  \log p_{\theta}(x)  
   &= KL(q_{\phi}(z|x)||p_{\theta}(z|x)) + L(\theta, \phi; x)
 \end{aligned}
 \end{equation}
 where $KL(Q||P)$ is KL between two distributions Q and P, $L(\theta, \phi; x)$ is the ELBO which is defined as
\begin{equation}
L(\theta, \phi; x) = E_{q_{\phi}(z|x)} (- \log q_{\phi}(z|x) + \log p_{\theta}(x,z))
\end{equation}
The term $q_{\phi}(z|x)$ is to extract latent feature from the observed data $x$ and it is called encoder generally. By minimizing KL, we try to find $q_{\phi}(z|x)$ that can approximate the true posterior distribution $p_{\theta}(z|x)$. Because $L(\theta, \phi; x)$ is non-negative and $\log p(x)$ is fixed, then minimizing $KL(q_{\phi}(z|x)||p_{\theta}(z|x))$ is equivalent to maximizing  $L(\theta, \phi; x)$. We can rewrite $L(\theta, \phi; x)$ as
 
 \begin{equation}
 \label{equ:lower_bound_rewrite_1}
 \begin{aligned}
  L(\theta, \phi; x) = E_{q_{\phi}(z|x)} \log p_{\theta}(x|z) - KL (q_{\phi}(z|x)||p_{\theta}(z))
  \end{aligned}
 \end{equation}
where the first term in the RHS of Eq.(\ref{equ:lower_bound_rewrite_1}) is the reconstruction error (RE), and the second term in the RHS is the KL between the prior and the posterior. Those two values play different roles during the approximation. We will introduce them in details in the next section.

\subsection{VAE with KKT Condition}

In practice, we find that the RE is usually the main error, while the term of KL is regarded as the regularization to enforce $q_{\phi}(z |x)$ to be close to $p_{\theta}(z | x)$, which is relatively small. If we constrain the KL term into a small component $\epsilon$ to gain a tighter lower bound, the goal is transformed to maximize the RE, namely  $\max_{\theta,\phi}E_{q_{\phi}(z|x)}\log p_{\theta}(x|z)$~\cite{higgins2017beta}.
 Then the objective function is changed with the inequality constraint:
 \begin{equation}
 \label{equ:rewrite_low}
 \begin{aligned}
  & \max_{\theta}E_{q_{\phi}(z|x)} \log p_{\theta}(x|z) \\&\quad \quad \text{subject to} \quad KL(q_{\phi}(z|x)||p_{\theta}(z))< \epsilon
  \end{aligned}
 \end{equation}

Then it changes to the optimization problem with the inequality constraints which can be solved by KKT, since $\lambda$ and $\epsilon$ are the nonnegative values. 
Under the KKT condition, Eq.(\ref{equ:rewrite_low}) can be rewritten as follows:
 \begin{equation}
 \label{equ:rewritte_KKT1}
 \begin{aligned}
  & \hat{L}(\theta, \phi;x, \lambda) = E_{q_{\phi}(z|x)} \log p_{\theta}(x|z)\\&-\lambda (KL(q_{\phi}(z|x)||p_{\theta}(z))-\epsilon)
  \end{aligned}
 \end{equation}
where $\hat{L}$ is the energy free objective function which can be regarded as the convex optimization problem, and $\lambda > 0$ is the Lagrangian multiplier, which is used to penalize the deviation of the constraint $KL(q_{\phi}(z|x)||p_{\theta}(z)) \le \epsilon$. Given that $\lambda > 0$ and $\epsilon >0 $, we have
 \begin{equation}
 \label{equ:rewritte_KKT1_1}
 \begin{aligned}
  \hat{L}(\theta, \phi; x, \lambda) \geq E_{q_{\phi}(z|x)} \log p_{\theta}(x|z) -\lambda KL(q_{\phi}(z|x)||p_{\theta}(z))
  \end{aligned}
 \end{equation}
If $\lambda=1$, then Eq.(\ref{equ:rewritte_KKT1_1}) reduces to the original VAE problem that proposed by Kingma et al.~\cite{kingma2013auto}. However, if $0< \lambda <1 $, then $\hat{L}(\theta, \phi;x, \lambda)>L(\theta, \phi;x)$, which is closer to the target $\log p_{\theta}(x)$. This is just the mathematical description of the fact that the more information in the latent variable $z$, the tighter of the lower bound is. Through the KKT condition, a loose constraint over the decoder is introduced. Empirical results show that VAE with KKT condition performs better than original VAE. Thus, in this paper, we use VAE with KKT condition as our basic model.

\subsection{Semi-supervised VAE}
When there is label information $y$ in the observed data, it is easy to extend Eq.(\ref{equ:rewritte_KKT1_1}) to include label information as follows~\cite{kingma2014semi}.
  \begin{equation}
 \label{equ:labeled_original}
 \begin{aligned}
  & \hat{L}(\theta, \phi; x, y, \lambda) \geq E_{q_{\phi}(z|x, y)} \log p_{\theta}(x|z, y)\\&+ \lambda (\log p(y)+\log p(z) - \log q_{\phi}(z|x,y))
  \end{aligned}
 \end{equation}
To achieve the semi-supervised learning, Kingma et al.~\cite{kingma2014semi} introduce a classifier $q_{\phi}(y | x)$ to Eq.(\ref{equ:labeled_original}), which results in
  \begin{equation}
 \label{equ:unlabel_posterior}
 \begin{aligned}
  {U}(\theta, \phi, y; x, \lambda) = \sum_{y}q_{\phi}(y|x)\hat{L}(\theta, \phi; x, y, \lambda) +H(q_{\phi}(y|x))
  \end{aligned}
 \end{equation}
 Apart from the Eq.(\ref{equ:labeled_original}) and Eq.(\ref{equ:unlabel_posterior}), the classification loss over the label information $E_{p(x, y)}\log q_{\phi}(y|x)$ is added into the objective function when facing with the labeled data. However, in this paper, the discriminative information is added from scratch and an equality constrained VAE is proposed, in order to highlight the contribution of labeled data.

\section{The Proposed Framework}
\label{sec:model_des}
In this section, we introduce the details of the proposed framework. Instead of using a classifier to incorporate the label information, we seek to directly use label information to regularize the latent representation so as to reduce the number of parameters.

\subsection{Disentangled Representation}
In order to incorporate the label information to the latent representation, we assume that the latent representation can be divided into two parts, i.e., the disentangled variable and non-interpretable variable. The disentangled variable captures the categorical information, which can be used for prediction task. Therefore, we can use label information to constrain the disentangled variable. The non-interpretable variable can be a vector comprised of any dimensions that combine other uncertain information from the data. For the simplicity of notation, we use $u$ to denote the non-interpretable representation and $v$ to denote the disentangled variable. With $u$ and $v$, the encoder can be rewritten as $q_{\phi}(u, v|x)$. We further assume that the disentangled variable and the non-interpretable variable are independent condition on $x$, i.e.,
\begin{equation}
  q_{\phi}(u, v|x) = q_{\phi}(u | x) q_{\phi} (v | x)
\end{equation}

It is a reasonable assumption because given $x$, the categorical information is only dependent on $x$ and $v$, which captures the categorical information, is independent of $u$ given $x$. This means that there is seldom information about the category information in $u$, which is validated in the experiment part.

 Now $q_{\phi}(u|x)$ is the encoder for the non-interpretable representation, and $q_{\phi}(v|x)$ is the encoder for the disentangled representation. Based on those assumptions, Eq.(\ref{equ:labeled_original}) is written as:
 \begin{equation}
 \label{equ:ecvae}
 \begin{aligned}
   & \hat{L}(\theta, \phi; x, \lambda) \geq E_{q_{\phi}(u|x),q_{\phi}(v|x)} \log p_{\theta}(x|u, v)\\&+ \lambda (\log p(v)+\log p(u) - \log q_{\phi}(u|x) - \log q_{\phi}(v|x))\\
   &=RE_{(u,v)} - \lambda (KL_{u}+KL_{v})
 \end{aligned}
 \end{equation}
where $RE_{(u,v)} = E_{q_{\phi}(u|x),q_{\phi}(v|x)} \log p_{\theta}(x|u, v)$, which represents the RE given the variables $(u,v)$. $KL_{u}$ and $KL_{v}$ denote the $KL(q_{\phi}(u|x)||p(u))$ and $KL(q_{\phi}(v|x)||p(v))$ respectively. From the above equation, we can see that the categorical information is extracted from the data, i.e., captured in disentangled variable $v$. Now if we have partial labels given, we can directly use the label information to regularize $v$.

With $v$ capturing the categorical information, there are many ways to regularize $v$. Inspired by the work of~\cite{higgins2017beta}, we add equality constraint on $v$ over the ELBO, where the equality constraint is to enforce the disentangled representation $v$ to be close to the label information $y$. In this work, we consider two ways to add the constraint over the ELBO as discussed below.

\subsection{SDVAE-I}
The first way we consider is the cross entropy between $y$ and $v$, i.e.,
\begin{equation}
  U = \sum_{i}^{|y|} y_{i}\log q_{\phi}(v_{i}|x)
\end{equation}
where $y$ is the observed one-hot label coding, and $|y|$ is the category number, $q_{\phi}(\cdot)$ is encoder for the disentangled variable $v$. This is a popular loss function for supervised learning and does not introduce any new parameters. Therefore, we choose this as the loss function for regularizing the disentangled variable $v$. We name this method Semi-supervised Disentangled VAE I (SDVAE-I). By adding this loss function to Eq.(\ref{equ:ecvae}), the objective function of SDVAE-I is given as:
 \begin{equation}
 \label{equ:objective_function_ecvae1}
 \begin{aligned}
  & \hat{L}(\theta, \phi; x, \lambda, \mu)\geq RE_{(u,v)} -\lambda (KL_{u}+KL_{v}) + \mu U
  \end{aligned}
 \end{equation}
where $\mu$ is the weight parameter to control the contribution of $U$. For the unlabeled data, the equality condition will be $U=0$.

\subsection{SDVAE-II}
 The drawback of the SDVAE-I is obvious because the training results depend on the number of the labeled data heavily.  However, for semi-supervised learning, there is usually a small size of the labeled data available. Thus, it is hard for the disentangled variable to capture the category information. To this end, inspired by the idea in~\cite{xu2017variational}, we use REINFORCE process to learn latent information as the equality constraint. The idea is as follows. First, it is well known that a tight ELBO means a better estimation for the posterior, which means that a bigger ELBO gives a better approximation. To make the posterior distribution inference heuristic, we treat the ELBO as reward $R$ in the reinforcement learning, and the encoder $q_{\phi}(v|x)$, which is to approximate the posterior distribution, is treated as the policy networks. During the inference, this process is depicted as followed,
 \begin{equation*}
 \label{equ:reinforce1}
 J(x,v;\phi) = \sum_{i}^{|v|} R \cdot q_{\phi}(v_{i}|x) 
 \end{equation*}
  Where $R$ denotes the $ RE_{(u,v)} - (KL_{u}+KL_{v})$. To reduce the variance during the Monte Carlo estimation for the reinforce estimation, a careful selected $c$ is subtracted from the reward $R$~\cite{mnih2014neural}, in this paper, $c$ is the mean value of the disentangled value $v$. To use the label $y$ explicitly, the function $f(y)$ is added as the factor before the reward $R$. When facing the unlabeled data $f(y)=1$ to slack their influence, and it is $f(y)=y$ to the labeled data to reinforce the label signal. Then we add this part to the objective function directly as the equality constraint. Its update rule is showed in Eq.(\ref{equ:updates}). 
\begin{equation}
\label{equ:updates}
\Delta \phi = f(y)(R-c)\nabla_{\phi} \log q_{\phi}(v|x)
\end{equation}

  The reward will lead the model to find its way. The disentangled variable $v$ acts as the classifier and helps the model distinguish the latent information, and this lets agent get a better prediction.

 Also, the term log-likelihood expectation on disentangled variable $v$ is added as the information entropy, which will be calculated both in labeled data and the unlabeled data. This not only helps to reduce the large variance of the disentangled information but also lets the model be guided by the labeled data and avoids the misconvergence during the training. Then the objective function in Eq.(\ref{equ:objective_function_ecvae1}) is changed into Eq.(\ref{equ:objective_function_ecvae2}). 

\begin{equation}
\label{equ:objective_function_ecvae2}
 \begin{aligned}
  & \hat{L}(\theta, \phi; x, \lambda, c, \beta_{1}, \beta_{2})\geq RE_{(u,v)} -\lambda(KL_{u}+KL_{v}) \\& + (\beta_{1} R-c)\log q_{\phi}(v|x) + \beta_{2}H (q_{\phi}(v|x))
  \end{aligned}
\end{equation}
where $y$ is the label information, $\beta_{1}$ and $\beta_{2}$ are the coefficient parameters, and we name this model SDVAE-II. This model is guided by the labeled data, also learns from the unlabeled data using the reinforcement learning.

\subsection{With Inverse Autoregressive Flow}
A flexible inference usually is built from the normalizing flow, which is to give a better approximation for the posterior distribution~\cite{rezende2015variational}. The normalizing flow is the probability density transformation through a sequence of invertible mapping, and after the sequence processing, a valid probability distribution is obtained. Among them, the IAF ~\cite{kingma2016improved} is an effective normalizing flow, which can handle high-dimensional latent space.

In this paper, because the two different latent variables are extracted from the data directly, to make the posterior inference more flexible and enhance the ability in disentangled representation in high-dimension space, the IAF~\cite{yang2017improved} is applied in SDVAE-I and SDVAE-II. The chain is initialized with the output $\pi_{0}$ and $\delta_{0}$ from the encoder. Together with the random sample $\varepsilon \sim N(0, I) $, the non-interpretable variable $u$ is calculated as $u_{0} = \pi_{0}+\delta_{0} \otimes \varepsilon$. The way to update IAF chain is the same as that in the Long Short-Term Memory (LSTM) shown in Eq.(\ref{equ:iaf}).
 \begin{equation}
 \label{equ:iaf}
  u_{t} = \delta_{t} \otimes u_{t-1} + \pi_{t}
 \end{equation}
where $(\delta_{t}, \pi_{t})$ are the outputs of the auto-regression neural networks, whose input is the last latent variable $u_{t-1}$, and $t$ is the flow length.

\subsection{Training of SDVAE}
The models can be trained end-to-end using mini-batch with the ADAM optimizer~\cite{kingma2014adam}. The training algorithm is summarized in Algorithm \ref{algorithm:2}. In Line 1, we initialize the parameters. From Line 3 to Line 5, we sample a mini-batch to encode the input data as $u$ and $v$. From Line 6 to Line 10, we apply IAF. We then update the parameters from Line 11 to Line 13.
  
 \begin{algorithm}[ht]
\caption{Training algorithm of the proposed models.}
\begin{algorithmic}[1]
\STATE{Initialize the parameters $\epsilon, \phi, \theta$}
\REPEAT
\STATE{$x$ $\leftarrow$ Sample a mini-batch from the datapoints}
\STATE{$\epsilon \leftarrow$ Random sample from the noise distribution}
\STATE{$u, v\leftarrow q_{\phi}(u,v|x, \epsilon)$}
\IF{\textit{IAF}}
\FOR{$t<T$}
\STATE{$\hat{u} \leftarrow iaf(u,\theta)$}
\ENDFOR{}
\ENDIF{}
\STATE{$\hat{x}\leftarrow p_{\theta}(x|\hat{u},v)$ }
\STATE{$g\leftarrow \bigtriangledown_{\theta, \phi}\hat{L}(\theta, \phi; x, \lambda, c)$  \\
\small{Calculate the gradients of Eq.(\ref{equ:objective_function_ecvae2}}) for SDVAE-II, and Eq.(\ref{equ:objective_function_ecvae1}) for SDVAE-I.}
\STATE{$(\theta,\phi)\leftarrow $Update with gradients $g$}
\UNTIL{model convergence}
\end{algorithmic}
\label{algorithm:2}
\end{algorithm}

\subsection{Discussion}
In this subsection, we will discuss the differences between the previous works~\cite{kingma2014semi,xu2017variational} and our work. 

Firstly, the assumptions are different. In this work, we assume that the non-interpretable variable $u$ and disentangled variable $v$ at the same time are from both the labeled data and the unlabeled data. Furthermore, we assume that these two variables are independent. However, it is not the same in the previous works, they only extract the latent variable $u$ from the data. When there is no label information, label variable $y$ inferred from the $x$ with the shared parameters from $q_{\phi}(u|x)$ or inferred from $u$ directly.

Then, based on different assumptions, there are differences between the previous works in mathematics. The ELBO with two independent latent variable inferences is written as Eq.(\ref{equ:ecvae}), and it is different from that in Eq.(\ref{equ:labeled_original}) who only has one latent variable $u$ inference. Furthermore, if we ignore the assumption difference, when facing with the labeled data in previous works, their objective function is a special case in Eq.(\ref{equ:objective_function_ecvae2}) when $\beta_{1}=\beta_{2}=0$.

When the label is missing, previous works apply the marginal posterior inference over the label information which is shown in Eq.(\ref{equ:unlabel_posterior}). In this paper, it is the inference for both latent variable inference over the $u$ and $v$, and this is shown in Eq.(\ref{equ:unlabed_proposed}).
 \begin{equation}
 \begin{aligned}
 \label{equ:unlabed_proposed}
 U(x) = RE_{(u,v)}-\lambda (KL_{u}+KL_{v})+\beta_{2}H(q_{\phi}(v|x))
 \end{aligned}
\end{equation}

\section{Experimental Results}
\label{sec:exp}
In this section, we conduct experiments to validate the effectiveness of the proposed framework. Specifically, we want to answer the following questions: (1) Is the disentangled representation able to capture the categorical information? (2) Is the non-interpretable variable helpful for the data reconstruction? (3) Is the proposed framework effective for semi-supervised learning? To answer the above questions, we conduct experiments on image and text datasets, respectively.

\subsection{Baseline Models}
To make fully evaluation for SDVAE, and make sure the comparisons cover a broad range of semi-supervised learning methods, the following baselines are involved in the comparisons.
\begin{itemize}
  \item \textbf{K-Nearest Neighbors (KNN)} KNN~\cite{peterson2009k} is the traditional graph-based method applied in classification.

  \item \textbf{Support Vector Machine (SVM)} SVMs have been widely used in classification works, and Transductive SVM (TSVM)~\cite{joachims2006transductive} is a classic discriminative model in semi-supervised learning. Apart from TSVM, SVM can also be applied in text data analysis, so the model NBSVM~\cite{wang2012baselines} is compared in the text data. 

  \item \textbf{CNNs} CNNs~\cite{krizhevsky2012imagenet} have been widely used in the image data and the text data. The model seq2-bown-CNN~\cite{johnson2014effective} is compared in the text data, and CNN itself is compared in the image data, respectively.

  \item \textbf{LSTMs} In the text data analysis, there are lots of semi-supervised learning methods by applying LSTM~\cite{hochreiter1997long}, models like LM-LSTM, SA-LSTM~\cite{dai2015semi}, etc., are compared in the text data.

  \item \textbf{Semi VAEs} Since VAE proposed~\cite{kingma2013auto}, it has been widely used in different fields, and semi-VAE~\cite{kingma2014semi} is popular in semi-supervised learning for the image data, and SSVAE is for the text data~\cite{xu2017variational} by applying the LSTM encode and decode the word embedding~\cite{li2018word}. Because of the proposed framework is a kind of VAE, the comparisons are made between those two frameworks.

\end{itemize}

Apart from the models mentioned above, models like RBM, Bow etc.~\cite{li2017learning,maas2011learning,li2018word}, are also included in the evaluation of the text data.

\subsection{Experiments on Image Datasets}
\subsubsection{Datasets Description} For image domain, we choose two widely used benchmark datasets for evaluating the effectiveness of SDVAE, i.e., MNIST~\cite{lecun1998gradient} and SVHN~\cite{netzer2011reading}. In the MNIST, there are 55,000 data samples in the train set and 10,000 data samples in the test set. In the SVHN, there are 73,257 data samples in the train set, and 26,032 data samples in the test set. Both datasets contain 10 categories. Before feeding the SVHN dataset into the model, the preprocessing of PCA is done.

\begin{table*}[htp]
\centering
\small
\caption{The classification errors on the MNIST data with part of labeled data, the number in brackets are the standard deviations of the results.}
\label{tab:res_mnist}
\begin{tabular}{ll|l|l|l|l}
\hline
 Models                         &                &  100          & 600      & 1000    & 3000     \\\hline
 KNN     &  \multirow{9}{*}{(\cite{kingma2014semi})} &25.81   &  11.44           &  10.07          &  6.04  \\
 CNN                           &                   &22.98        &  7.68            &  6.45           &  3.35  \\
 TSVM                          &                   &16.81        &  6.16            &  5.38           &  3.45  \\
 Semi-VAE(M1)+TSVM             &     &11.82($\pm$(0.25))      & 5.72($\pm$0.05)            &  4.24($\pm$0.07)          & 3.49($\pm$0.04) \\
 Semi-VAE(M2)                  &     &11.97($\pm$(1.71))                                & 4.94($\pm$0.13)             &  3.60($\pm$0.56)          & 3.92($\pm$0.63) \\
 Semi-VAE(M1+M2)               &       & 3.33($\pm$(0.14))                              & 2.59($\pm$0.05)             &  2.40($\pm$0.02)          & 2.18($\pm$0.04) \\\hline
 SDVAE-I                       &       & 5.49($\pm$(0.12))                    & {2.75($\pm$0.11)}    & {2.42($\pm$0.08)}  & {1.70($\pm$0.09)} \\
 SDVAE-II                      &        &    3.60($\pm$(0.06))              & {2.49($\pm$0.10)}    & {1.96($\pm$0.09)}  & {1.58($\pm$0.09)} \\
  SDVAE-I\&IAF                  &     & 3.33($\pm$0.03)                     & {2.74($\pm$0.06)}    & {2.24($\pm$0.08)}  & {1.33($\pm$0.09)} \\
 SDVAE-II\&IAF                 &       & 2.71($\pm$(0.32))    &  1.97($\pm$0.14)     & 1.29($\pm$0.11)   &  1.00($\pm$0.05)  \\
 \hline
\end{tabular}
\end{table*}

\subsubsection{Model Structure}
For the image data, the encoder is a deep network composed of two convolutional layers followed by two fully connected layers. The convolutional layers are used to extract features from the images while the fully connected layers are used to convert the features to the non-interpretable variable and the disentangled variable. The decoder is a network composed of two fully connected layers to map the latent features back to images. Dropout~\cite{srivastava2014dropout} is applied to both the encoder and decoder networks.

\subsubsection{Analysis on Disentangled Representation}
The first experiment is to explore how the non-interpretable variable $u$ and disentangled variable $v$ perform in the image reconstruction. The experiment is conducted on the MNIST dataset. In the training data, we randomly select 3000 data samples as labeled data and the remaining samples are unlabeled. The dimension of the disentangled variable $v$ is 10 which is the same as the category number, and the label information can be got from disentangled variable $v$ directly. And the dimension of $u$ is 50.

\begin{figure}[H]
\centering
\begin{minipage}[t]{0.44\linewidth}
\centering
\includegraphics[width=1.0\linewidth]{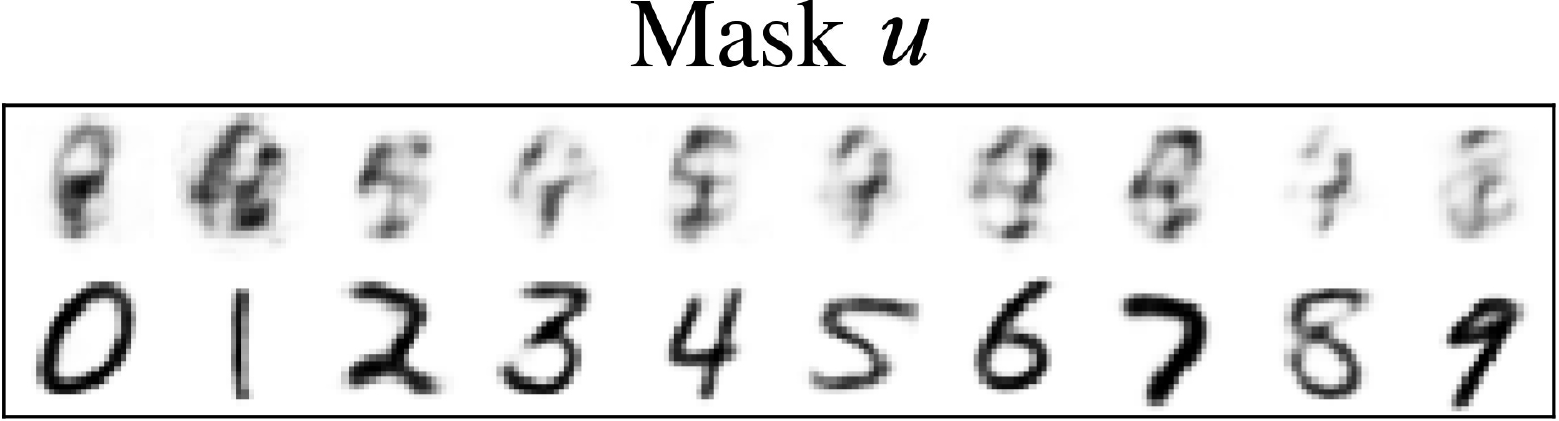}
\end{minipage}
\begin{minipage}[t]{0.44\linewidth}
\centering
\includegraphics[width=1.0\linewidth]{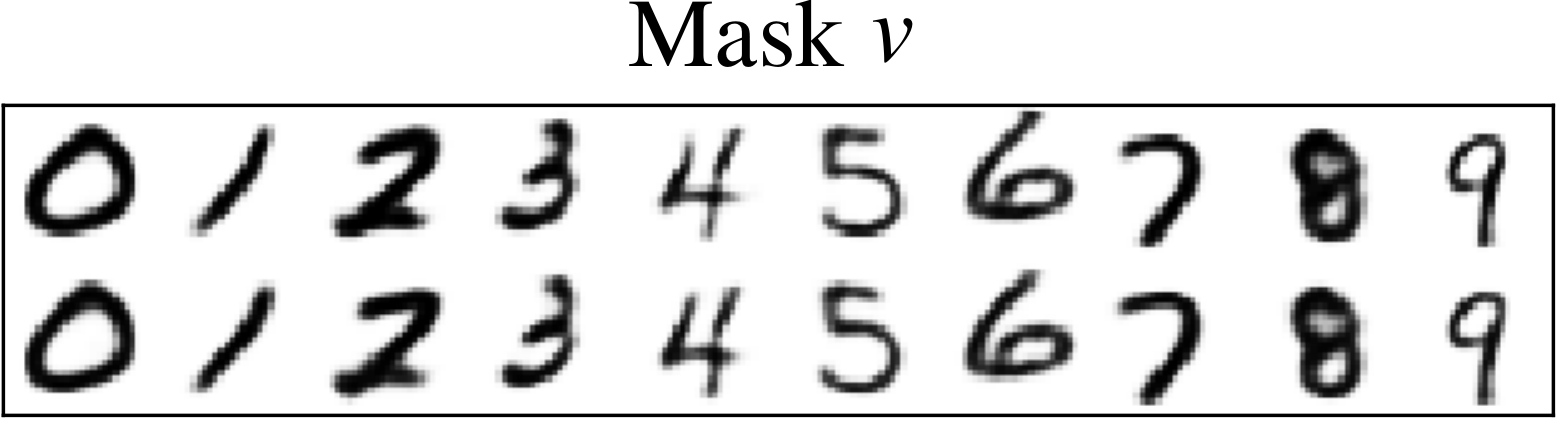}
\end{minipage}
\caption{The first row in left figure and the right figure are the reconstruction images with the variable $u$ and variable $v$ masked respectively, and the images in the second row in both figures are the test images original.}
  \label{fig:mnist_masked}
\end{figure}

\begin{figure}[H]
\centering
\begin{minipage}[t]{0.45\linewidth}
\centering
\includegraphics[width=1.0\linewidth]{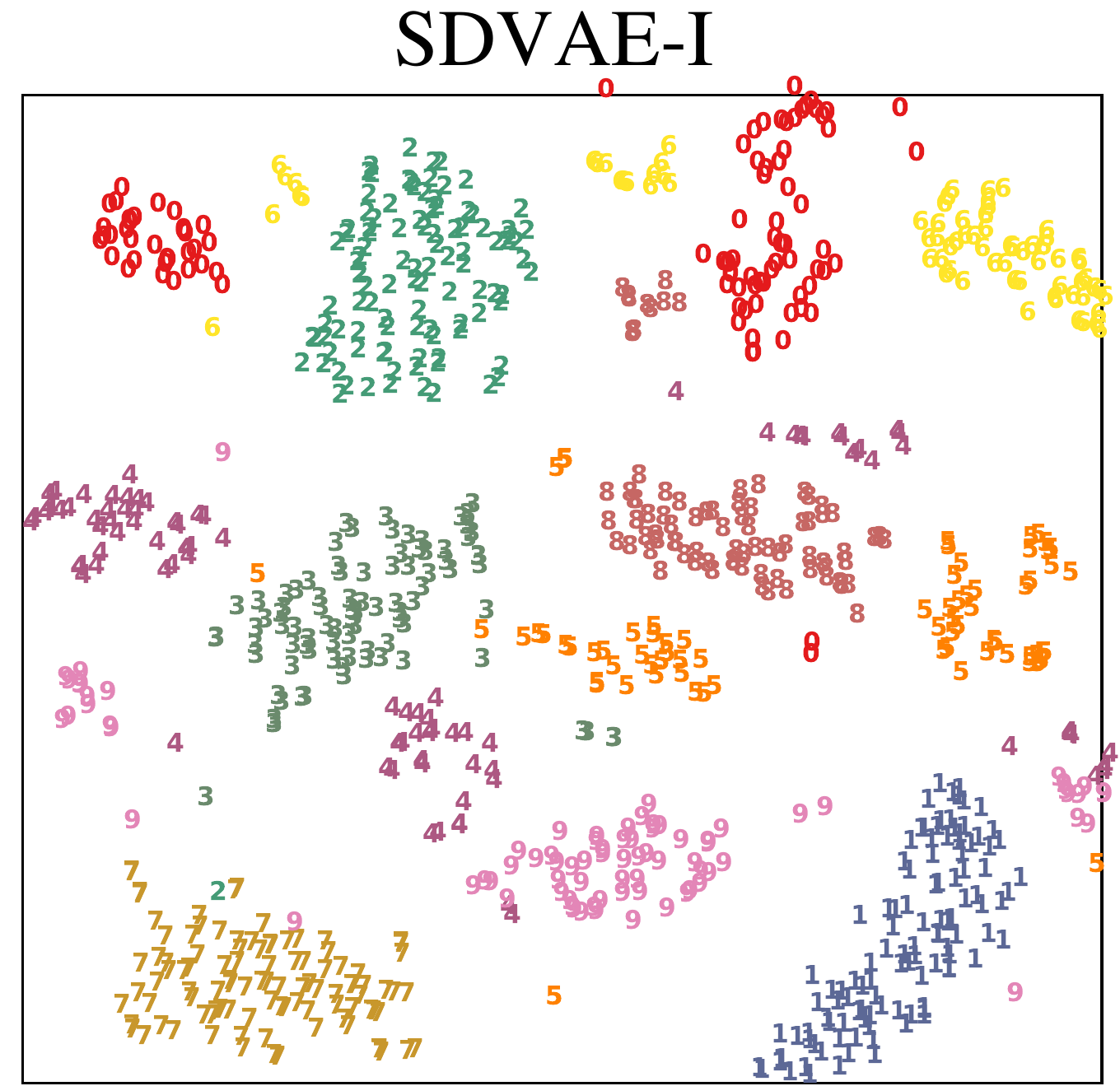}
\end{minipage}
\begin{minipage}[t]{0.43\linewidth}
\centering
\includegraphics[width=1.0\linewidth]{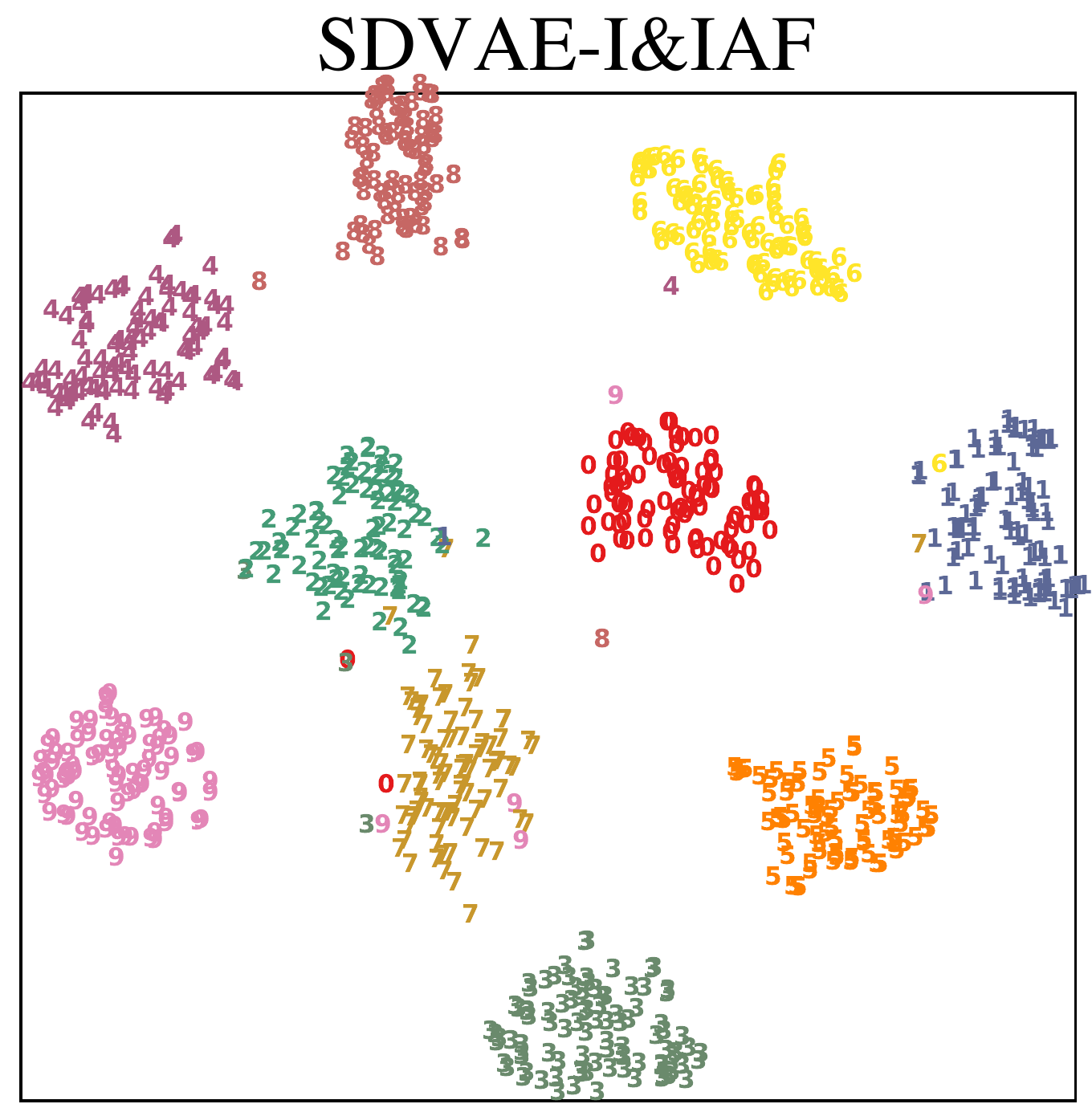}
\end{minipage}
\begin{minipage}[t]{0.44\linewidth}
\centering
\includegraphics[width=1.0\linewidth]{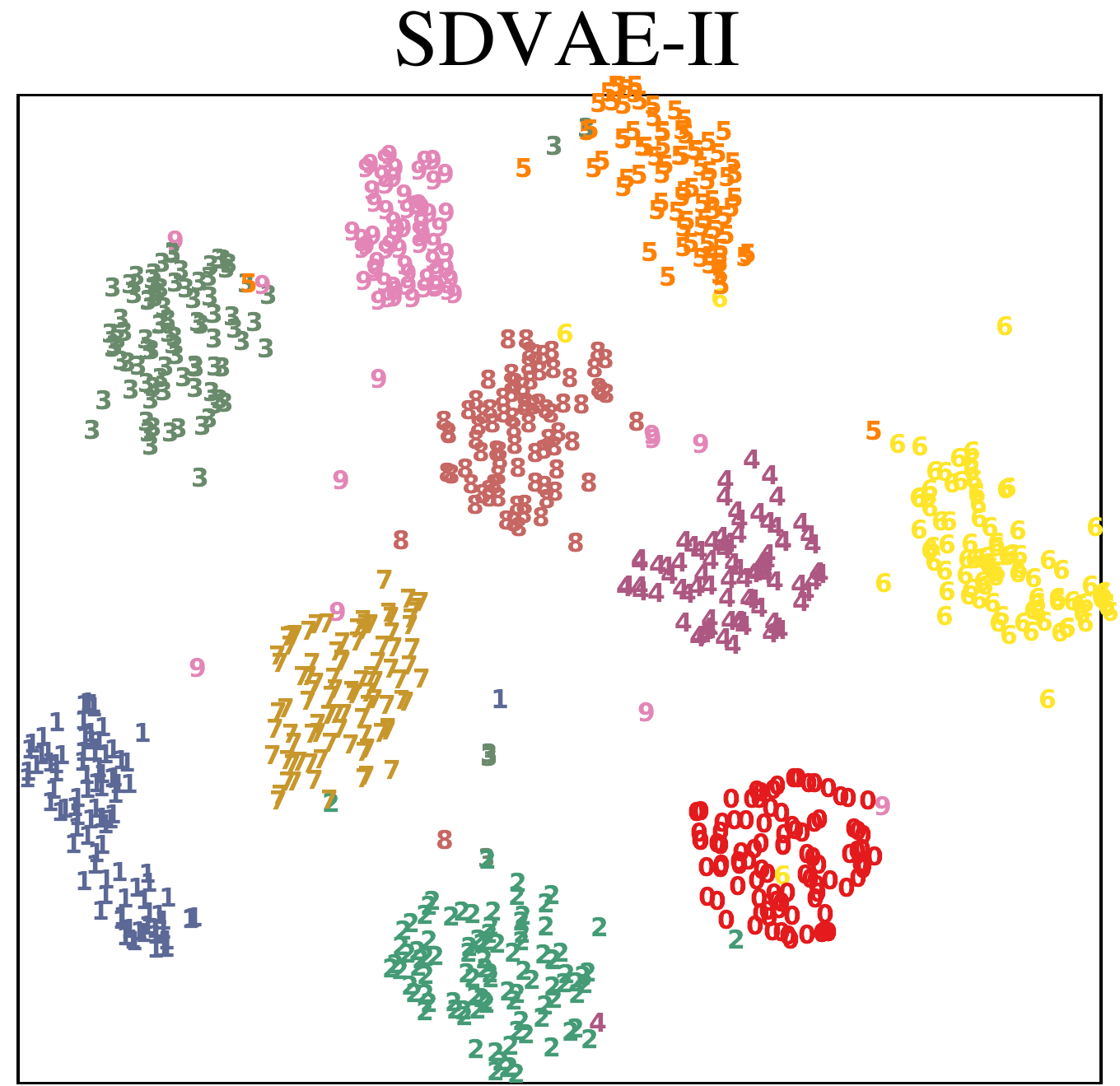}
\end{minipage}
\begin{minipage}[t]{0.445\linewidth}
\centering
\includegraphics[width=1.0\linewidth]{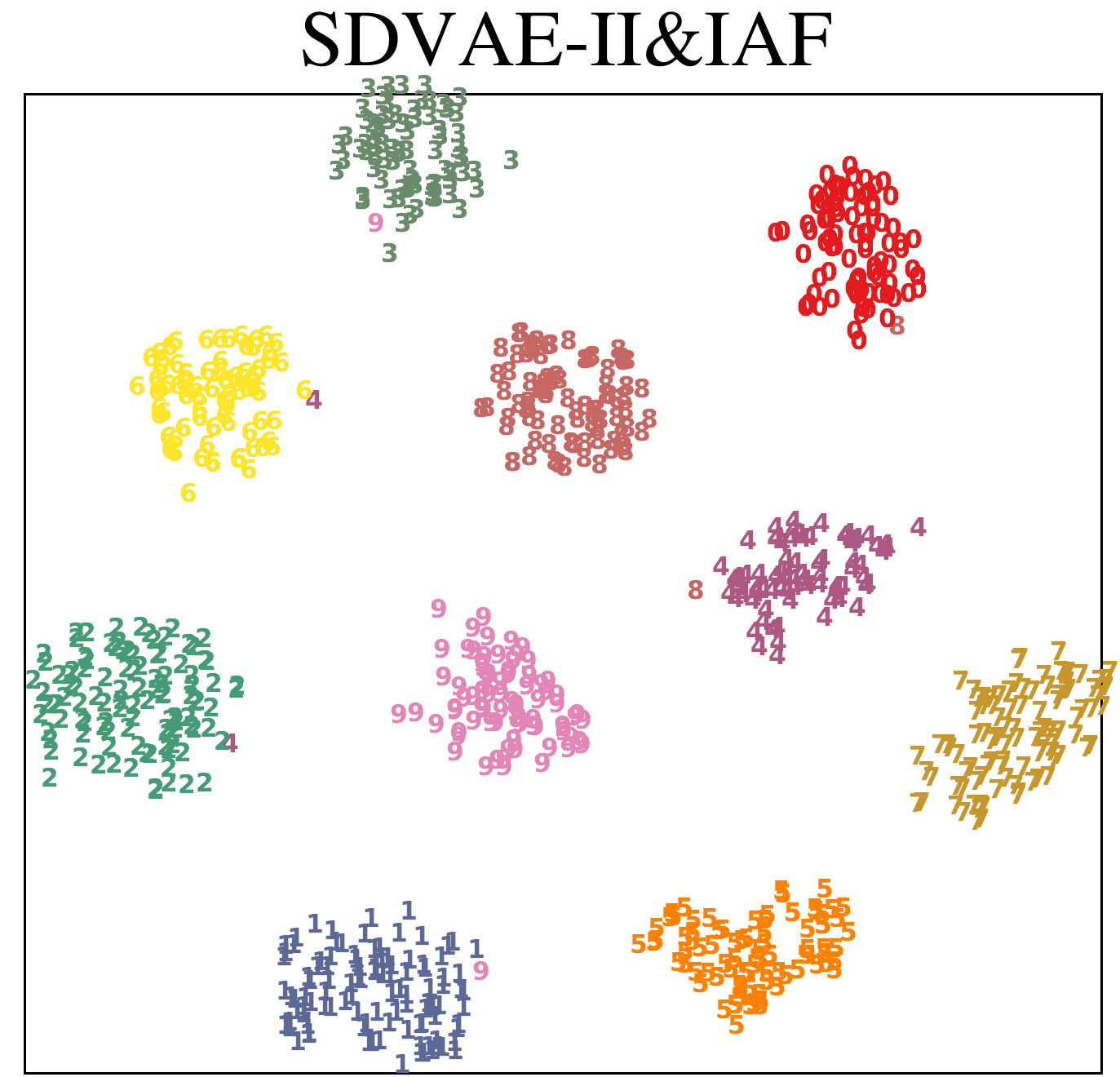}
\end{minipage}
\caption{The t-SNE distribution of the latent variable $v$ from proposed models, and different categories are in different colors with number.}
  \label{fig:mnist_tsne}
\end{figure}

We first train the model to learn the parameters. Then we use the trained model to learn latent representation on the test data. After learning the representations, we mask $u$ and $v$ in turn to see how they affect the reconstruction of input image. Two sample results are shown in Fig.\ref{fig:mnist_masked}. We also use t-SNE~\cite{van2008visualizing} to visualize $v$ of the testing data. The results from those four models (SDVAE-I, SDVAE-I\&IAF, SDVAE-II and SDVAE-II\&IAF) are shown in Fig.\ref{fig:mnist_tsne}.

From Fig.~\ref{fig:mnist_masked} and Fig.~\ref{fig:mnist_tsne}, we can see that the disentangled variable $v$ mainly captures the categorical information, and it has little influence over the reconstruction task, because when variable $u$ is masked, the reconstructed pictures are blurred. More specifically, from Fig.~\ref{fig:mnist_tsne}, we can see that images of the same class are clustered together, implying that the disentangled variable $v$ captures the categorical information. In addition, we find that cluster SDVAE-I gives the worst visualization as clusters have intersections, while SDVAE-I\&IAF and SDVAE-II\&IAF give better visualization, which suggests that SDVAE-I\&IAF and SDVAE-II\&IAF are better at capturing the categorical information, and the bounds of clusters in SDVAE-II are also clear enough.

\begin{figure}[H]
\centering
\begin{minipage}[t]{0.235\linewidth}
\centering
\includegraphics[width=1.0\linewidth]{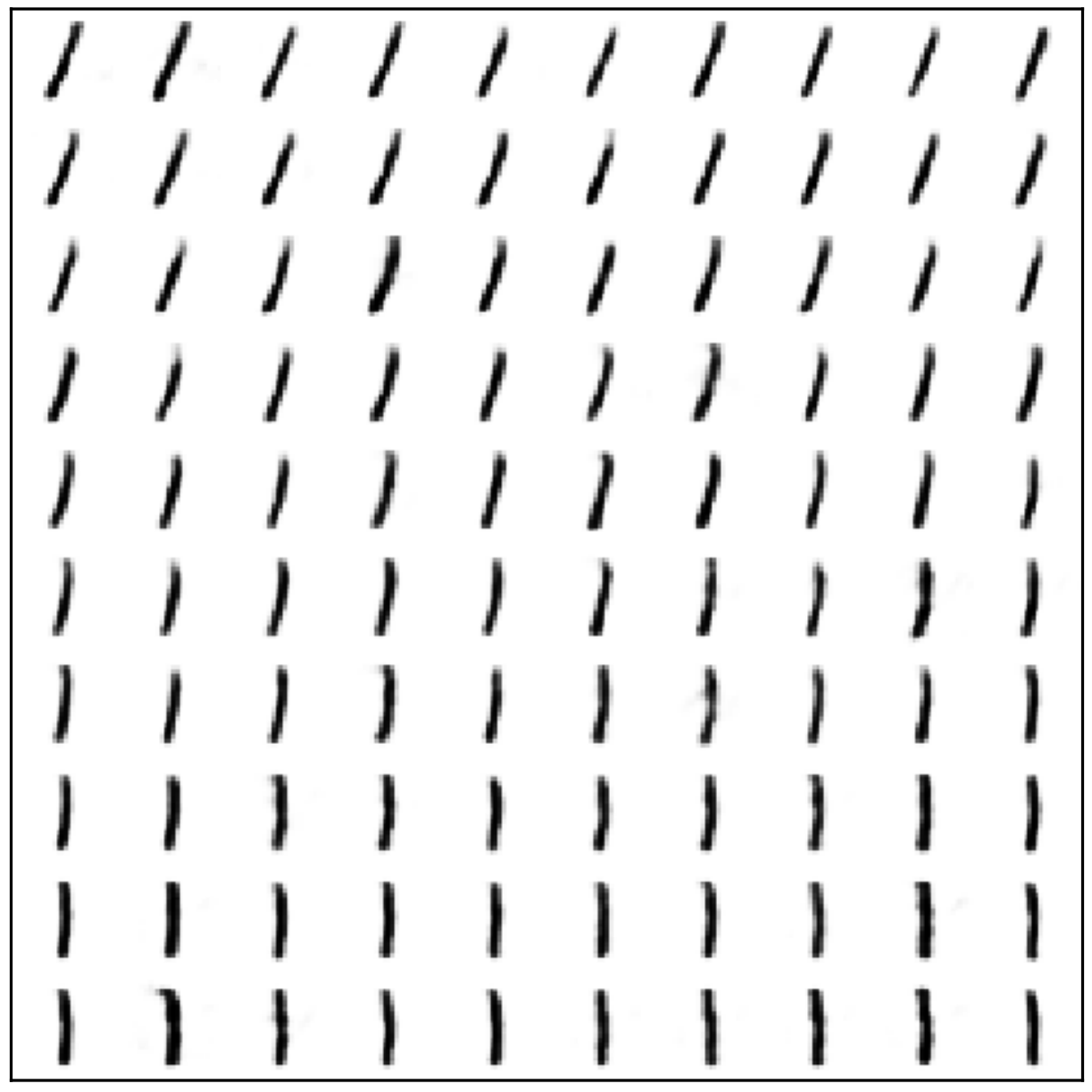}
\end{minipage}
\begin{minipage}[t]{0.235\linewidth}
\centering
\includegraphics[width=1.0\linewidth]{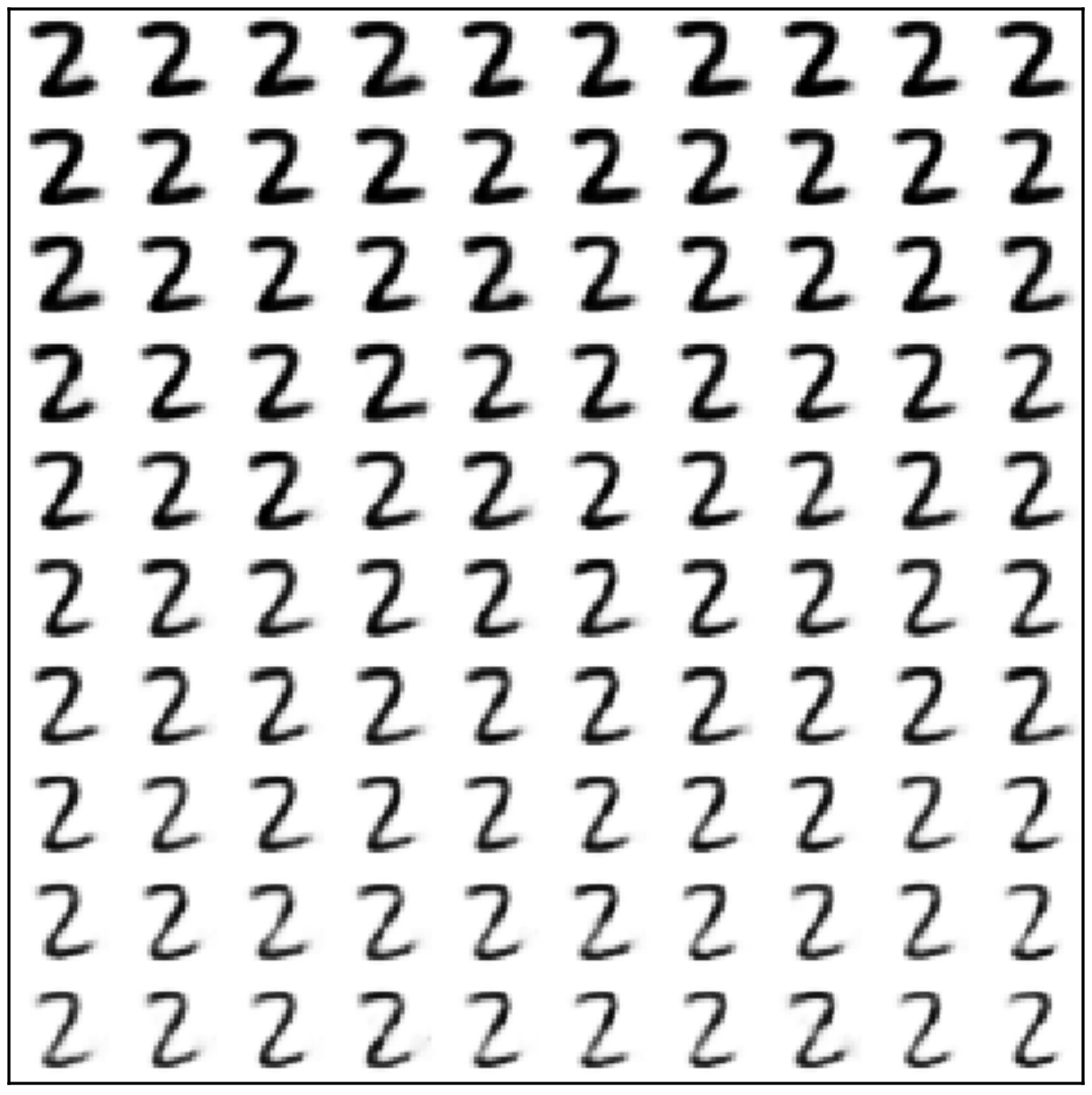}
\end{minipage}
\begin{minipage}[t]{0.235\linewidth}
\centering
\includegraphics[width=1.0\linewidth]{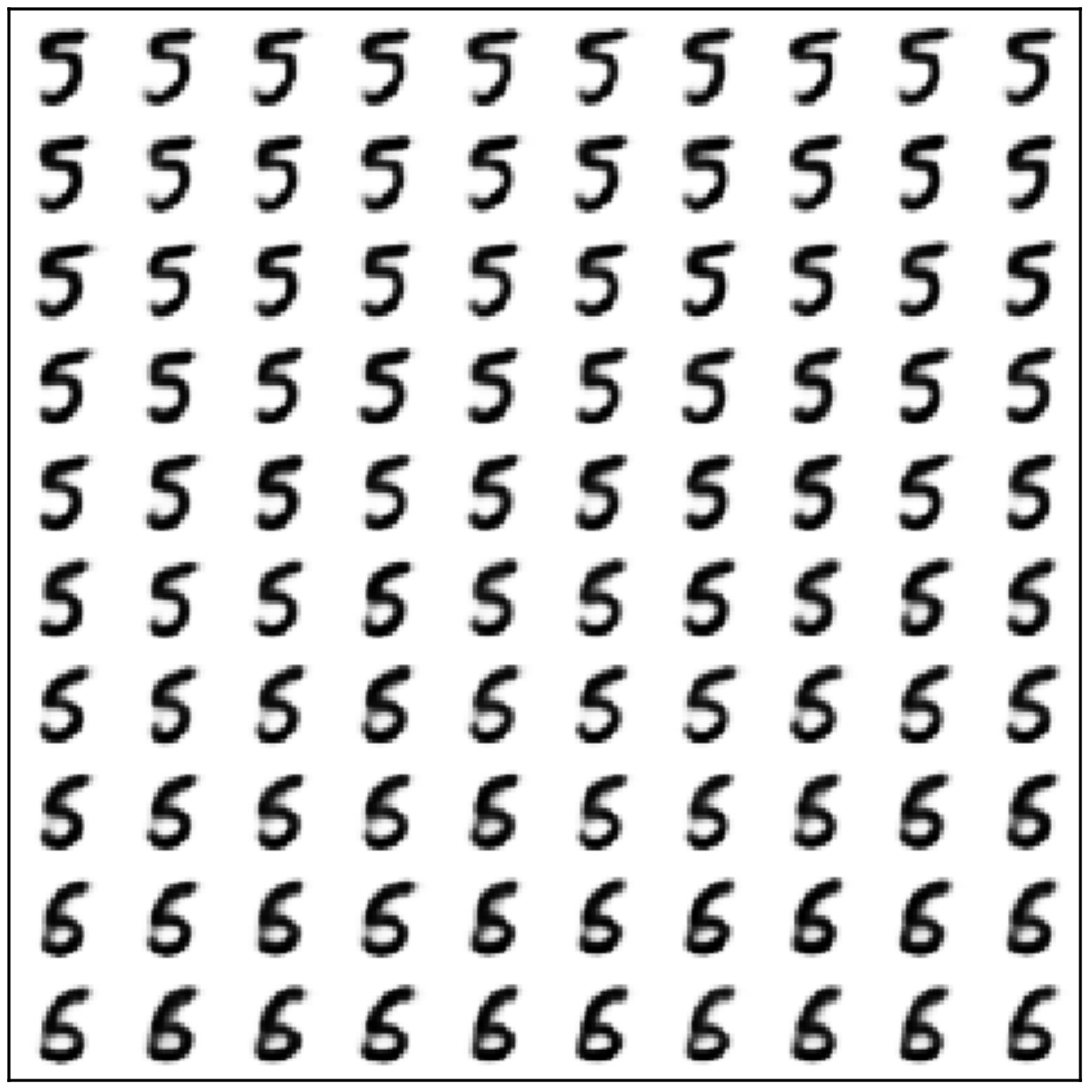}
\end{minipage}
\begin{minipage}[t]{0.235\linewidth}
\centering
\includegraphics[width=1.0\linewidth]{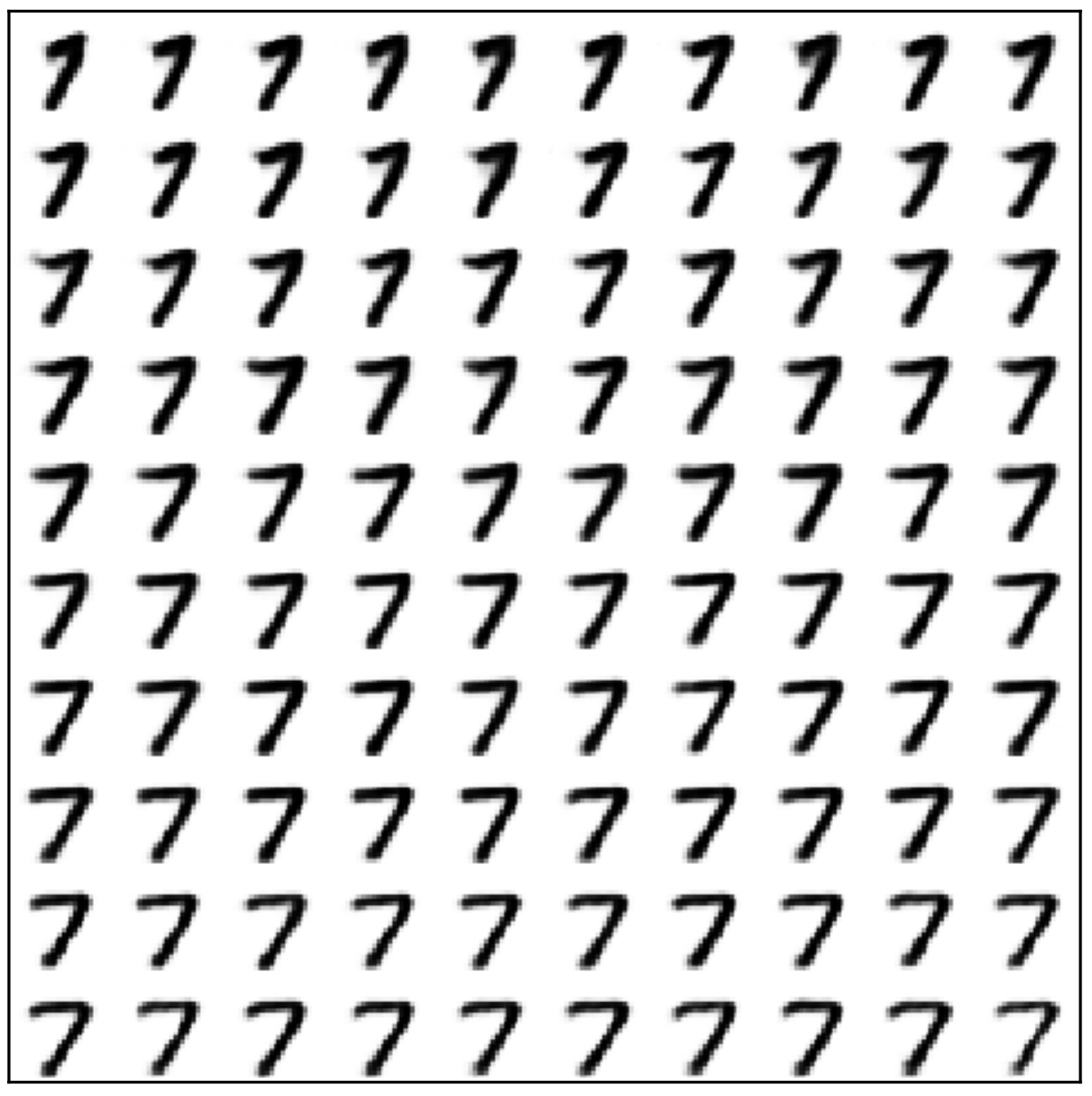}
\end{minipage}
\caption{The reconstruction images by varying $u$ in a certain dimension.}
  \label{fig:mnist_cevae1}
\end{figure}

From Fig.~\ref{fig:mnist_masked}, we can see that when $v$ is masked, $u$ still reconstructs the input image well, indicating that $u$ is appropriate for reconstruction. To explore how variable $u$ takes effect in the image reconstruction, we range a certain dimension of $u$ from -2 to 2 on the specific labeled image, and the selected results are shown in Fig.~\ref{fig:mnist_cevae1}.
From the image, we can see that $u$ can control different properties in image reconstruction with different dimensions, such as italic controlling, bold controlling, transform control, and the style controlling, etc. These can be seen from images in Fig.~\ref{fig:mnist_cevae1} left to right.

\subsubsection{Semi-Supervised Learning}
Furthermore, we conduct experiments to test the proposed models in semi-supervised learning on MNIST. We randomly select $x$ points from the training set as labeled data, where $x$ is varied as $\{100, 600, 1000, 3000\}$. The rest training data are used as unlabeled data. We compare with state-of-the-art supervised and semi-supervised classification algorithms, which are used in~\cite{kingma2014semi}. The experiments are conducted 10 times and the average accuracy with standard deviation are showed in Table~\ref{tab:res_mnist}. Note that the performances of the compared methods are from~\cite{kingma2014semi} too. From this table, we can see the proposed model SDVAE-II\&IAF performs best in classification and makes the least classification errors (in black bold format) with the small part of the labeled data. SDVAE-II also has better performance than previous models. Although SDVAE-I performs not as good as other proposed models, it still can achieve state-of-the-art results.

To further validate the observation, we also conduct the semi-supervised learning over the SVHN, another popularly used dataset. SVHN has 73,257 training samples and 26032 test samples. Among the training data, we randomly select 1000 data samples as labeled data and the rest as unlabeled data. The results are shown in Table~\ref{tab:results_of_svhn}. Similarly, among the model of same kinds, SDVAE-II performs better and SDVAE-II\&IAF gives the best performance both with the process of PCA preprocessing or without PCA preprocessing.

  \begin{table}[htp]
\centering
\small
\caption{The results on the SVHN data, the number in brackets are the standard deviations of the results.}
\label{tab:results_of_svhn}
\begin{tabular}{cc|c}
\hline
Method    &                                         & Test error rate          \\\hline
KNN       &\multirow{5}{*}{(\cite{kingma2014semi})} & 77.93\% ($\pm$0.08)    \\
TSVM      &                                         & 66.55\% ($\pm$0.10)   \\
Semi-VAE(M1)+KNN &                                  & 65.63\% ($\pm$0.15)   \\
Semi-VAE(M1)+TSVM&                                  & 54.33\% ($\pm$0.11)    \\
Semi-VAE(M1+M2)   &                                 & 36.02\% ($\pm$0.10)    \\ \hline
SDVAE-I  &      \multirow{3}{*}{Without}            & 47.32\%  ($\pm$0.13)    \\
SDVAE-II &     \multirow{3}{*}{Preprocessing}       & 44.16\%  ($\pm$0.14) \\
SDVAE-I\&IAF   &                                    & 46.92\%  ($\pm$0.12) \\
SDVAE-II\&IAF  &                                    & 34.25\%  ($\pm$0.13) \\ \hline
SDVAE-I  &       \multirow{3}{*}{With PCA}          & 33.68\%  ($\pm$0.11) \\
SDVAE-II &      \multirow{3}{*}{Preprocessing}      & 29.88\%  ($\pm$0.12) \\
SDVAE-I\&IAF   &                                    & 29.58\%  ($\pm$0.14) \\
SDVAE-II\&IAF  &                                    & 29.37\%  ($\pm$0.12) \\ 
\hline
\end{tabular}
\end{table}

\subsection{Experiments on Text Dataset}
\subsubsection{Dataset Description}
To test the model on text data, the IMDB data~\cite{maas2011learning} is used. This dataset contains 25,000 train samples and 25,000 test samples in two categories.

\subsubsection{Model Structure}
 In the application of the text data, the encoder is also the CNN, but unlike for image data, there are two CNNs parallelized together, which are referring from~\cite{kim2014convolutional}. One is extracting the feature at the word level, and the other is extracting the feature at the character level. As to the decoder, we applied the conditioned LSTM~\cite{wen2015semantically}, which is given as follows:
\begin{equation}
\begin{aligned}
&f_{t} = \sigma(W_{f}[u;v]+U_{f}h_{t-1}+b_{f})\\
&i_{t} = \sigma(W_{i}[u;v]+U_{i}h_{t-1}+b_{i})\\
&o_{t} = \sigma(W_{o}[u;v]+U_{o}h_{t-1}+b_{o})\\
&I_{t} = W_{c}[u;v]+U_{c}h_{t-1}+b_{c} \\
&c_{t} = f_{t}\otimes c_{t-1}+i_{t}\otimes \sigma(I_{t}) \\
&h_{t} = o_{t}\otimes relu(c_t)
\end{aligned}
\end{equation}
The conditional LSTM is same as the vanilla LSTM except for the current variable, which is replaced by the concatenation of the latent variable $u$ and $v$. The techniques of dropout~\cite{srivastava2014dropout} and batch normalization~\cite{ioffe2015batch} are both utilized in the encoder and decoder networks.

\subsubsection{Analysis on Disentangled Representation}
\begin{figure}[H]
  \begin{center}
  \subfigure[Unlabeled Data]{
  \label{fig:imdb_tsne_unlabel}
     \includegraphics[width=0.8\linewidth]{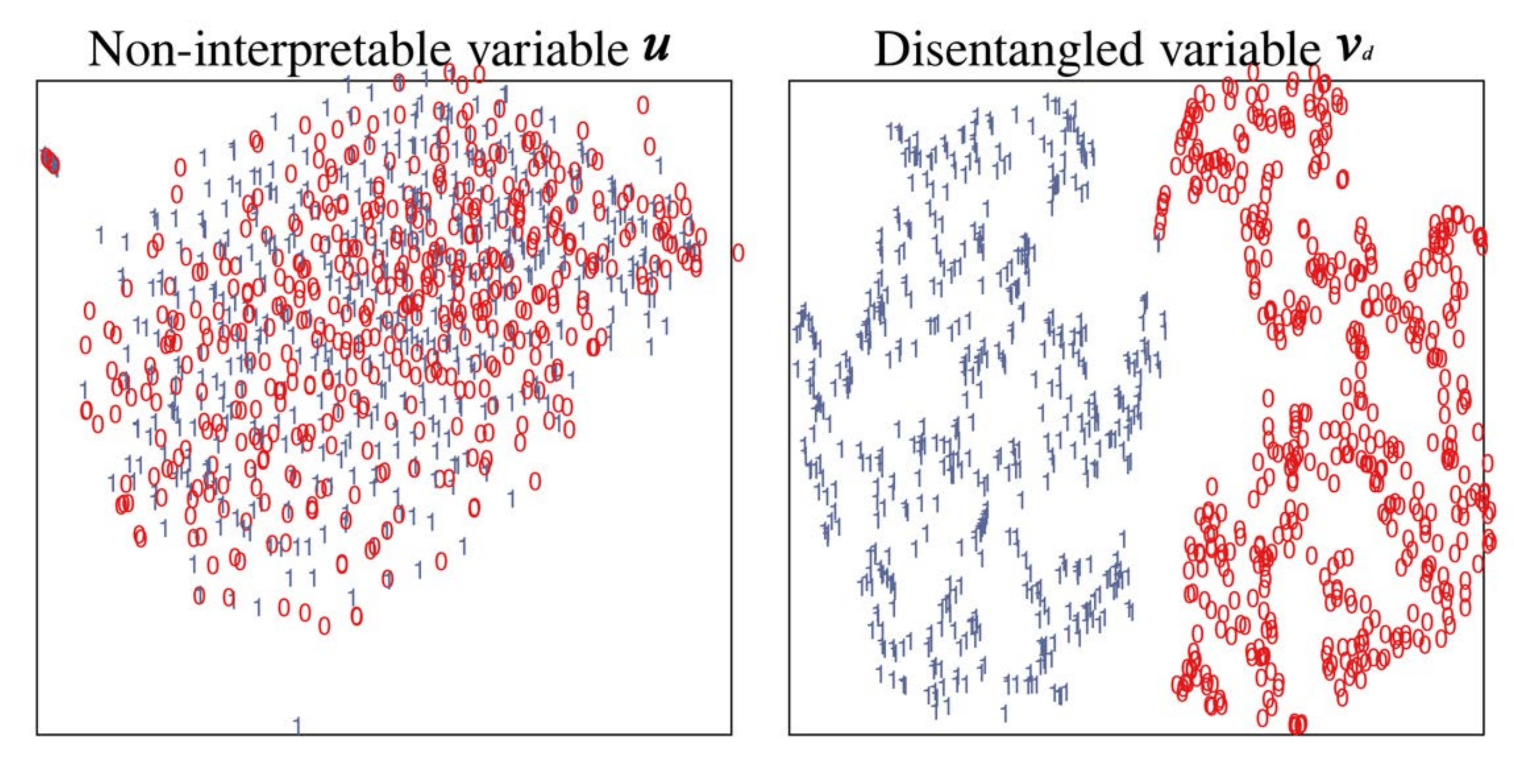}}\hfill
     \vskip -0.2em
  \subfigure[Test Data]{
  \label{fig:imdb_tsne_test}
  \includegraphics[width=0.8\linewidth]{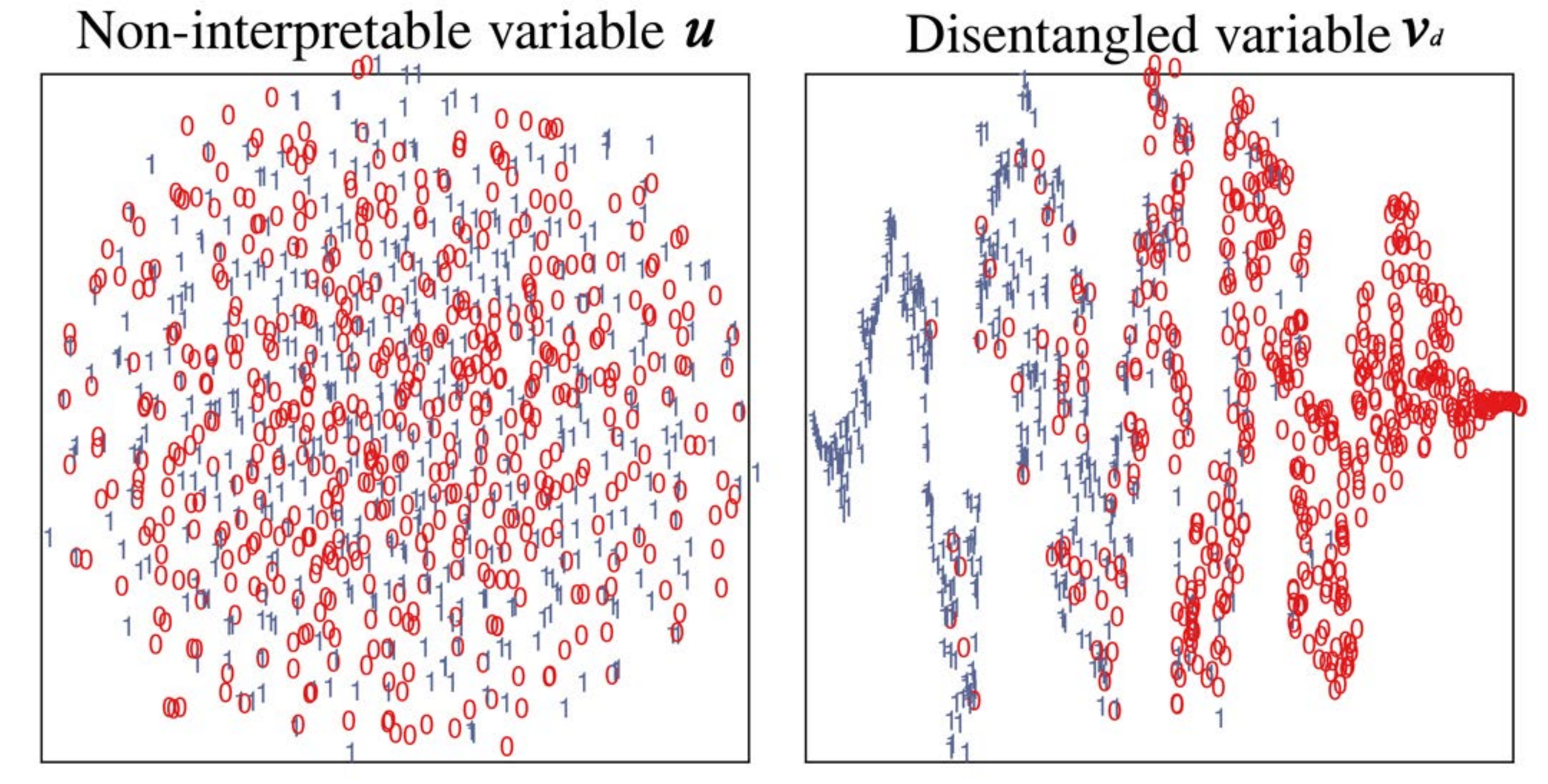}}
  \end{center}
    \vskip -1.8em
 \caption{The left figure is the t-SNE distribution of the non-interpretable variable $u$, the right figure is the t-SNE distribution of the disentangled variable $v$ correspondingly. Different categories are in different colors with number. }
  \label{fig:imdb_tsne1}
\end{figure}

 We randomly select 20K samples from the training set as the labeled data, and others are unlabeled during the training. Similarly, we use the t-SNE to visualize the disentangled variable $v\in N^{2}$ and the non-interpretable variable $u\in N^{50}$ from the proposed model on the test data and unlabeled data. Results are showed in Fig.\ref{fig:imdb_tsne1},

From the left figure in Fig.~\ref{fig:imdb_tsne1}, we can see that the disentangled representation $v$ can clearly separate the positive and negative samples while non-interpretable representation cannot points from two clusters are interleaved with each other. This suggests that the disentangled representation captures categorical information well, and there is seldom categorical information in the non-interpretable variable.

 \subsubsection{Semi-Supervised Learning}
We further conduct semi-supervised classification on the text dataset using the representation learned from previous experiments and fine tuning the model. Similarly, we compare with state-of-the-art semi-supervised learning algorithms. The average test error rate is reported in Table~\ref{tab:results_of_imdb}. From the results, we can see that: (i) SDVAE-II\&IAF outperforms the compared methods, which implies the effectiveness of the proposed framework for semi-supervised learning; and (ii) As we add reinforcement learning and IAF, the performance increases, which suggests the two components contribute to the model.
 \begin{table}[htp]
\centering
\small
\caption{The results on the IMDB data}
\label{tab:results_of_imdb}
\begin{tabular}{c|c}
\hline
Method                                       & Test error rate          \\\hline
LSTM (\cite{dai2015semi})                           & 13.50\%     \\
Full+Unlabeled+BoW  (\cite{maas2011learning})    & 11.11\%    \\
WRRBM+BoW       (\cite{maas2011learning})         & 10.77\%    \\
NBSVM-bi        (\cite{wang2012baselines})        & 8.78\%     \\
seq2-bown-CNN   (\cite{johnson2014effective})      & 7.67\%     \\
Paragraph Vectors (\cite{le2014distributed})       & 7.42\%    \\
LM-LSTM    (\cite{dai2015semi})                    & 7.64\%   \\
SA-LSTM   (\cite{dai2015semi})                     & 7.24\% \\
SSVAE-II\&LM (\cite{xu2017variational})            &7.23\% \\\hline
SDVAE-I                                             & 12.56\% \\
SDVAE-II                                            &7.37\% \\
SDVAE-I\&IAF                                         & 11.60\% \\
SDVAE-II\&IAF                                        &7.18\%  \\ 
\hline

\end{tabular}
\end{table}
 \vskip -1em

\subsection{Parameters Analysis}
There are several important parameters need to be tuned for the model, i.e., $\lambda$, $\beta_1$, $\beta_2$ and the length of IAF. In this section, we conduct experiments to analyze the sensitiveness of the model to the parameters.

\subsubsection{Effects of $\lambda$ and the IAF Length}

We firstly evaluate $\lambda$ and the length of the IAF chain, which are proposed in the works of $\beta$-VAE~\cite{higgins2017beta} and IAF~\cite{kingma2016improved}. These experiments are conducted over the MNIST training dataset. 

\begin{figure}[ht]
    \begin{center}
    \subfigure[Validate $\lambda$]{
        \label{fig:exp_lambda}
        \includegraphics[scale=0.25]{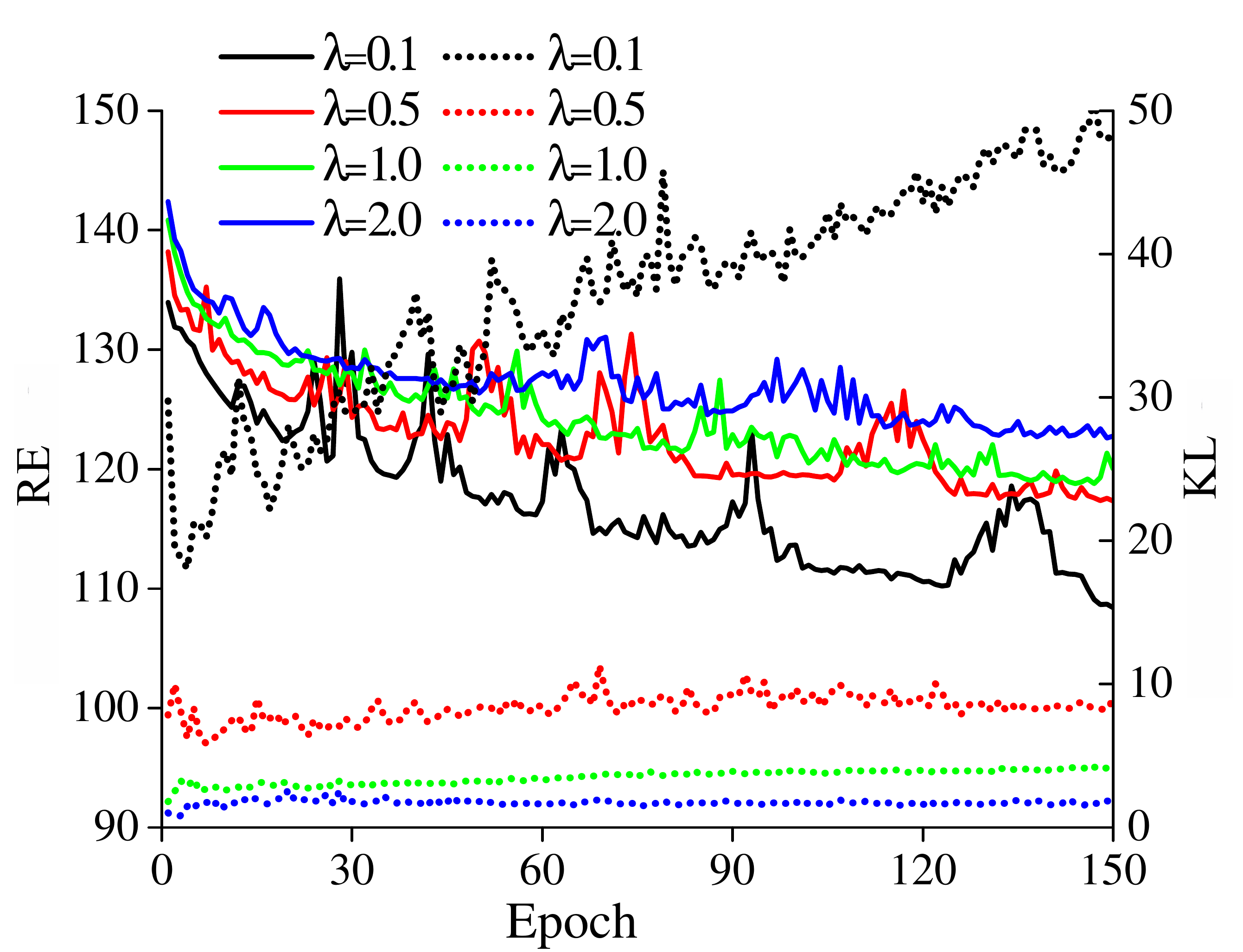}} 
    \subfigure[Validate Length of IAF]{
        \label{fig:exp_iaf}
        \includegraphics[scale=0.25]{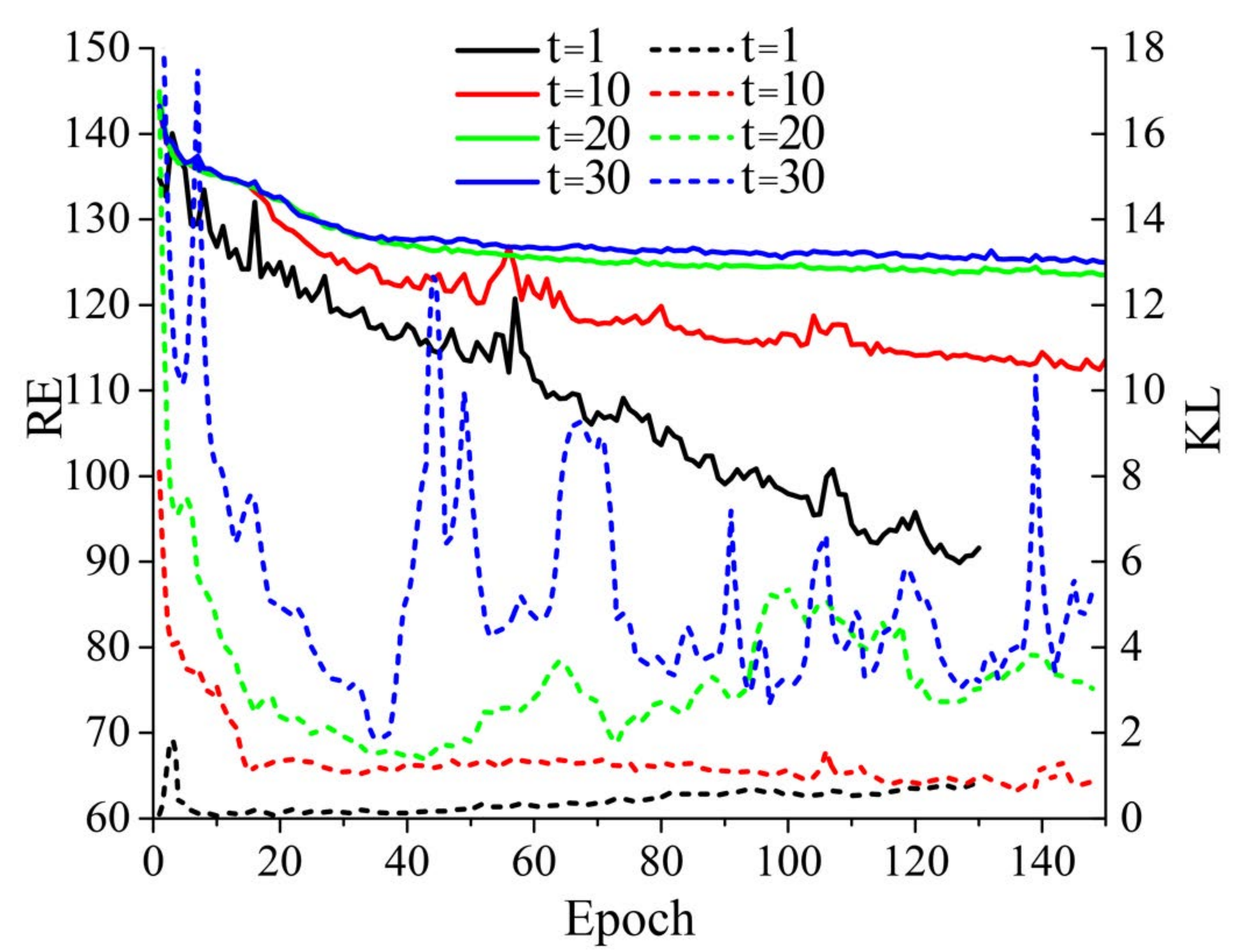}}
    \end{center}
    \vskip -1.5em
    \caption{The left y-axis in each figure is the RE which is axis of the solid lines, and the right y-axis is the KL which is axis of the dash lines.}
    \vskip -0.5em
\end{figure}

 For $\lambda$, we mainly focus on the objective function depicted in Eq.(\ref{equ:rewritte_KKT1_1}). Results with different $\lambda$ values are shown in Fig.~\ref{fig:exp_lambda}.
  From the results, we can see that it is better for $\lambda$ to have a small value, which not only leads to a rich information in the latent variable but also gets a better RE.
 But as described before, the large value of KL-divergence is also the cause of overfitting or the underfitting for the model.
 However, in the case of $\lambda=0.1$, there is a low RE, which is the sign of the good performance.
 
Then the model structure of the IAF chain is built according to the Eq.(\ref{equ:iaf}), and the results with different length are shown in the right figure in Fig.~\ref{fig:exp_iaf}. From the figure, we can see that it is not good to set the chain too long if it is a long IAF. The REs are not so good together with the KL, and the latent variable is very unstable. On the contrary, there is a stable increase in the KL, and a stable decrease RE when the length of the IAF chain is set to $1$. This means that under the good reconstruction, the latent variable captures more useful information. This is also validated in the results of the SDVAE-I\&IAF and SDVAE-II\&IAF. Thus, in the experiments about the IAF, its length is set to $1$ by default.

\begin{figure}[ht]
 \centering
 \includegraphics[scale=0.25]{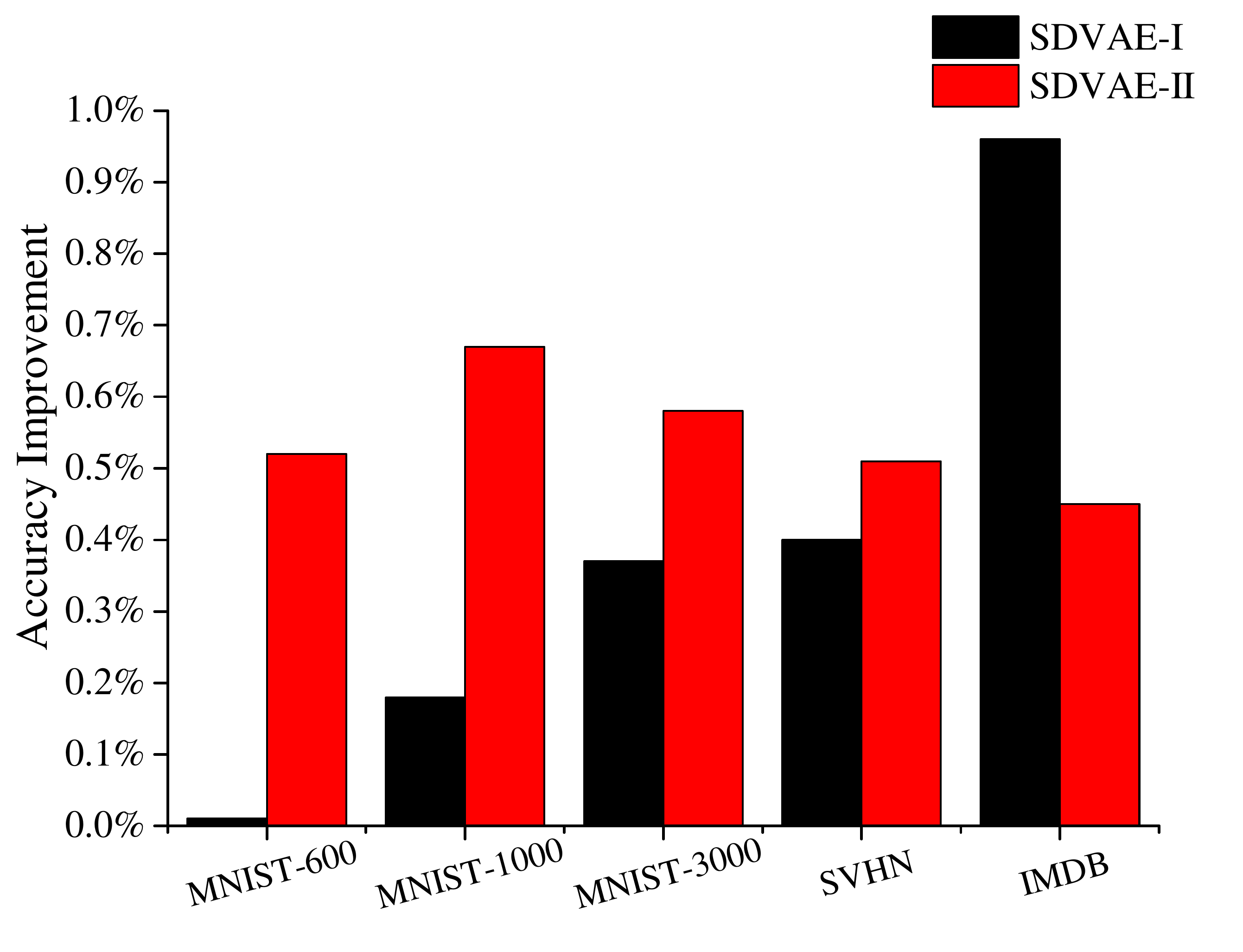} 
 \caption{The contribution of IAF in different datasets, and the value is the gap of accuracy from the SDVAE and SDVAE with IAF} 
 \label{fig:exp_iaf_contr} 
\end{figure}

We then get the improvement between the proposed model with IAF and that without IAF by subtracting the two performances, i.e., (${Accuracy}_{\text{\tiny SDVAE\&IAF}} - {Accuracy}_{\text{\tiny SDVAE}}$). And the results show in Fig.~\ref{fig:exp_iaf_contr}. From those results, we can see that the IAF is effective in feature learning, and there are improvements in every dataset used both in SDVAE-I and SDVAE-II. So, we can get the conclusion that it is useful to add IAF in proposed models both in text data and image data.

\subsubsection{Effects of $\beta_1$ and $\beta_2$}

\begin{figure}[htp]
    \begin{center}
    \subfigure[The Image data]{
        \label{fig:image_beta}
        \includegraphics[scale=0.247]{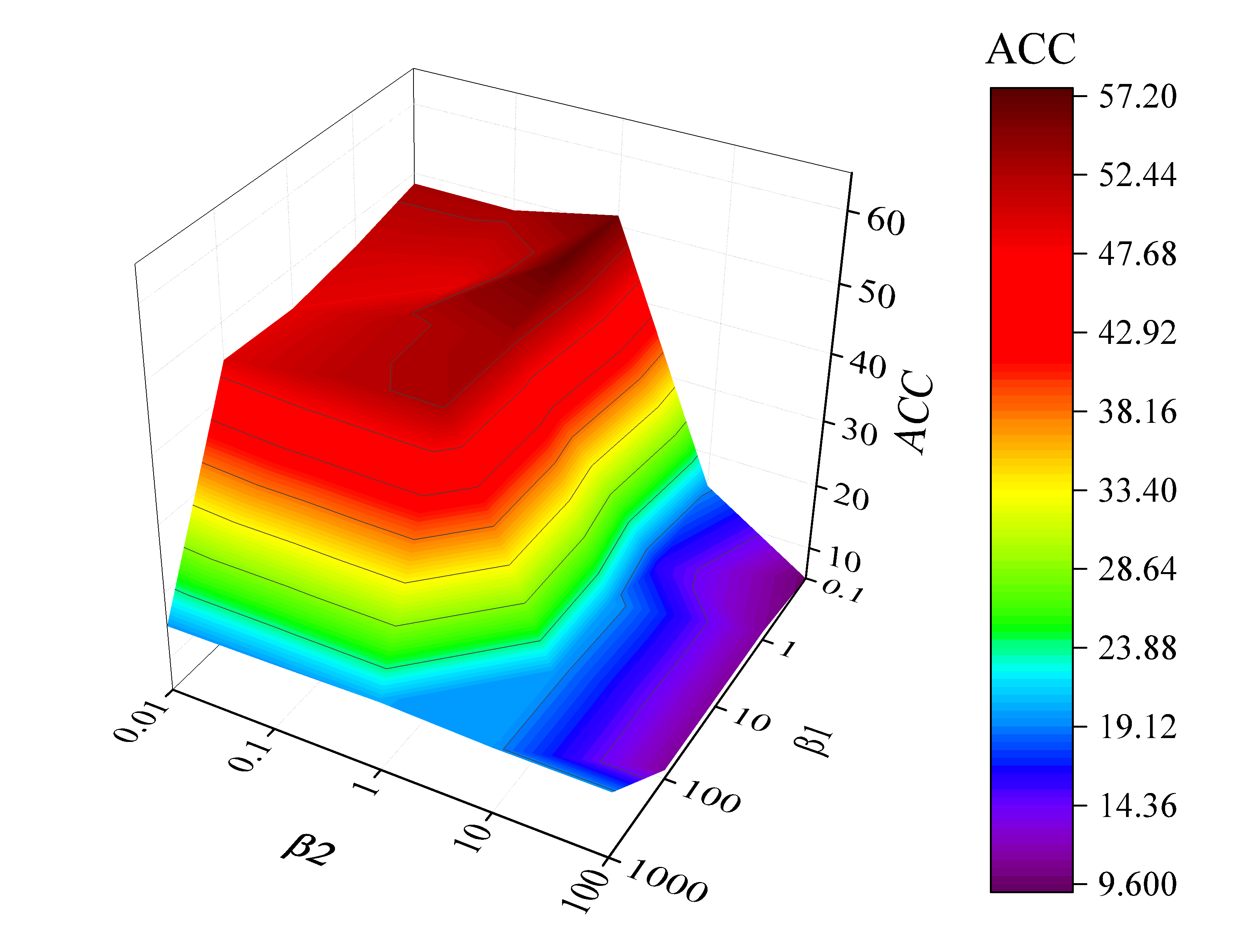}} \hfill
    \subfigure[The Text data]{
        \label{fig:text_beta}
        \includegraphics[scale=0.247]{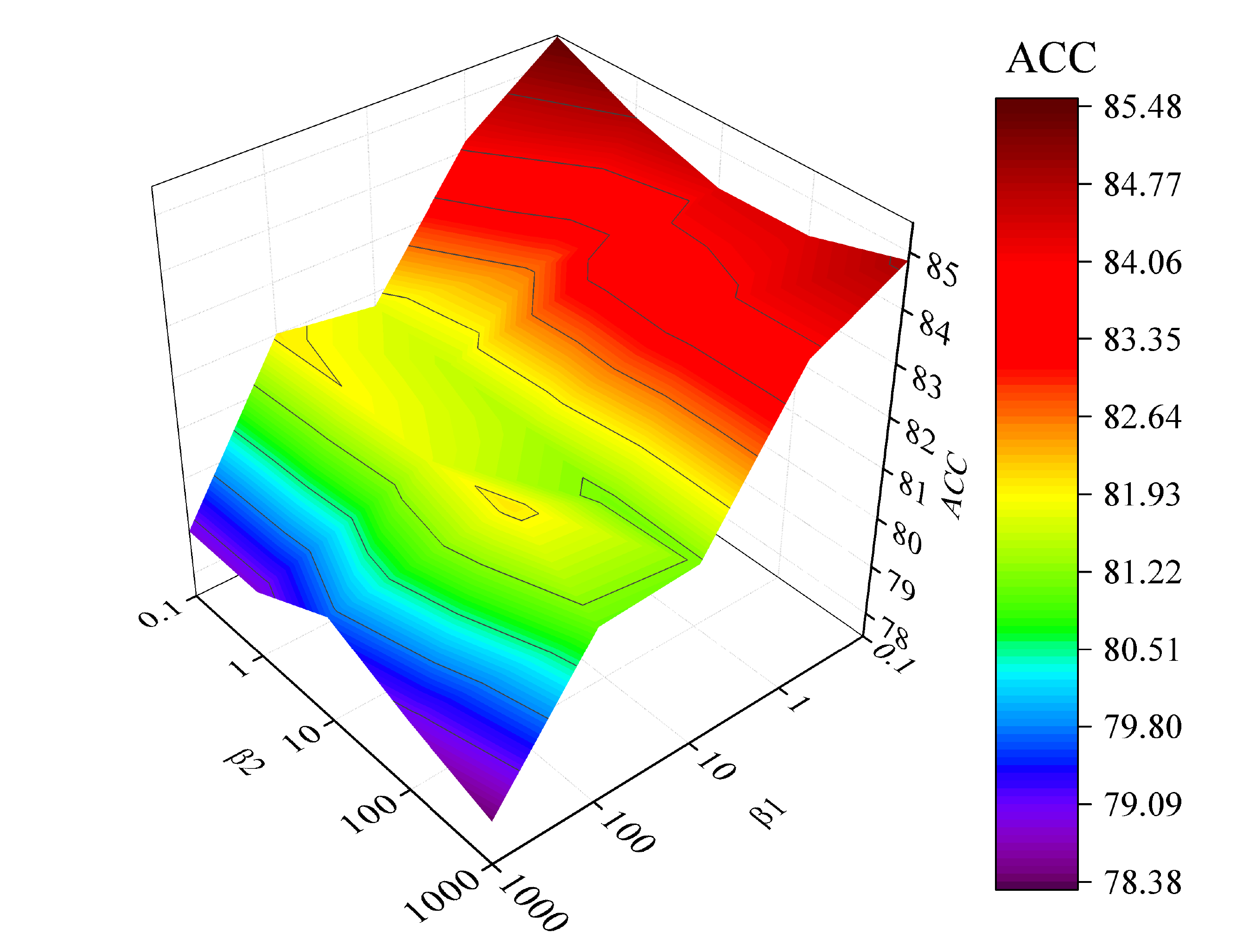}}
    \end{center}
    \vskip -1.5em
    \caption{The grid search results for the proper $\beta_1$ and $\beta_2$ finding.}
    \vskip -1em
\end{figure}

To decide the parameter $\beta_{1}$ and $\beta_{2}$ that in SDVAE-II, we made the grid search both on the text data and the image data. For the image data, the experiment is conducted on the SVHN dataset with 1000 labeled samples. Experimental results with $\beta_{1}$ ranges from $0.1$ to $1000$, and $\beta_2$ ranges from $0.01$ to $100$ are shown in Fig.\ref{fig:image_beta}. For the text data, the experiment is conducted on the IMDB data with 20,000 labeled samples. Experimental results with $\beta_{1}$ and $\beta_2$ range from $0.1$ to $1000$ are shown in Fig.\ref{fig:text_beta}.

From the Fig.~\ref{fig:image_beta}, we can see that, an acceptable range for $\beta_{1}$ in the image data is [0.1:100] and [0.01:10] for the $\beta_{2}$. Especially, when $\beta_{1}=0.1$ and $\beta_{2}=1$, it is achieving the best result.

For the text data, the results in the Fig.~\ref{fig:text_beta} show that the accuracy is not sensitive to $\beta_2$. However, when $\beta_1$ is small, the result will be more precise. In conclusion, it is better to set $\beta_1$ to 0.1 and $\beta_2$ can be set randomly.

\section{Conclusion} 
\label{sec:conclusions}
 In this work, we proposed models that extract the non-interpretable variable $u$ and the disentangled variable $v$ from data at the same time. The disentangled variable is designed to capture the category information and thus relieve the use of the classifier in semi-supervised learning. The non-interpretable variable is designed to reconstruct the data. Experiments showed that it could even reflect certain textual features, such as italic, bold, transform and style in the hand writing digital data during the reconstruction. These two variables cooperate well and each performs its own functions in the SDVAE. The IAF improves the model effectively on the basis of SDVAE-I and SDVAE-II. In particular, SDVAE-II\&IAF achieves the state-of-the-art results both in image data and the text data for the semi-supervised learning tasks.

 There are several interesting directions need further investigation. First, in this work, we choose CNNs as our encoder. We want to investigate more different deep neural networks to see which one gives the best performance under which condition. Second, in this work, we mainly focus on model reconstruction and semi-supervised classification. Another interesting direction is to study the generated images by generator learned by the proposed framework.


\section*{References}
\bibliography{myrefer}

\end{document}